\documentclass[11pt]{article}

\usepackage[preprint]{acl}
\usepackage[hidecomments]{nert}

\usepackage{times}
\usepackage{latexsym}

\usepackage[T1]{fontenc}

\usepackage[utf8]{inputenc}

\usepackage{microtype}

\usepackage{inconsolata}

\usepackage{graphicx}

\usepackage{tipa}
\usepackage{booktabs}
\usepackage{tcolorbox}

\urldef{\urlwikipedias}\url{https://meta.wikimedia.org/wiki/List_of_Wikipedias}
\urldef{\urlwiktionaries}\url{https://meta.wikimedia.org/wiki/Wiktionary#List_of_Wiktionaries}
\urldef{\urlconllueval}\url{https://universaldependencies.org/conll18/conll18_ud_eval.py}
\urldef{\urldictionaria}\url{https://dictionaria.clld.org/submit#best-practice}
\urldef{\urludgenrepage}\url{https://universaldependencies.org/contributing/genres.html}
\urldef{\urlopenrouter}\url{https://openrouter.ai/docs/guides/routing/provider-selection}

\newcommand{\lexicon}[0]{\texttt{LEX}}
\newcommand{\dep}[0]{\texttt{DEP}}

\newcommand{\conll}[0]{\texttt{CoNLLU}}
\newcommand{\syntax}[0]{\texttt{SYN}}
\newcommand{\all}[0]{\texttt{\lexicon+\syntax}}

\newcommand{\dev}[0]{\texttt{dev}}

\title{Out of Parameters, Saved by Syntax:\\\textsc{UD-PrMPT} for Parser Augmented Low Resource Machine Translation}

\author{
 \textbf{Emma Thronson} \quad
 \textbf{Abhishek Purushothama} \quad
 \textbf{Amir Zeldes}
\\
 Corpling Lab
 \\
 Georgetown University
 \\
\{\emldisplay{et726@georgetown.edu}{et726},
\emldisplay{ap2089@georgetown.edu}{ap2089},
\emldisplay{amir.zeldes@georgetown.edu}{amir.zeldes}\}\texttt{@georgetown.edu}
}

\begin{document}
\maketitle
\begin{abstract}
Recent work advancing machine translation quality using LLMs for very low resource languages has relied on the injection of lexicon-based prompt components to expose vocabulary to models. However, these approaches run into a ceiling when models must guess sentence meaning purely from plausible combinations of possible word senses mapped to strings. In this paper we test \textsc{UD-PrMPT}, an approach using syntactic analyses in the Universal Dependencies formalism as prompt components to improve MT in five diverse low resource languages: Beja, Chintang, Chukchi, Coptic, and Xibe. Our results show additive improvements which stack on top of the lexicon based approach which are stronger for higher quality manual analyses but consistent for predicted parses as well.
\end{abstract}

\section{Introduction}

\begin{figure*}[t!bh]
    \centering
    \includegraphics[width=1\textwidth]{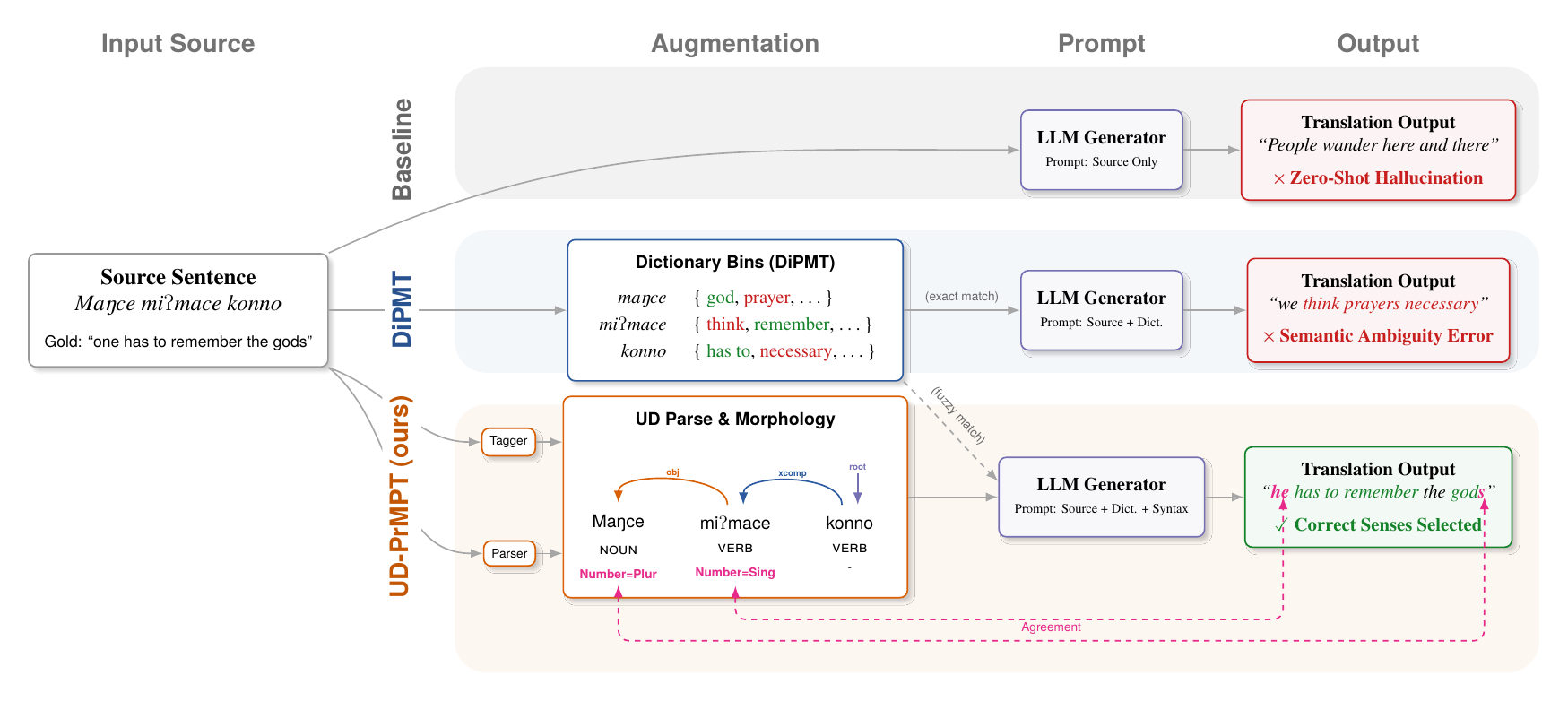}
    \caption{Chintang example: \textsc{DiPMT} prevents LLM translation hallucinations by proposing word senses, but these can be combined incorrectly. Our \textsc{UD-PrMPT} method integrates morphology tags and UD parses, showing `has to' is the main predicate and the noun cannot be the subject, and constraining agreement to produce better translations.}
    \label{fig:architecture}
    \vspace{-1em}
\end{figure*}

Recent advances in machine translation using LLMs in medium to high resource languages have benefited from models' inherent ability to condition on and serialize text in multiple languages using in-parameter linguistic capabilities \cite{robinson-etal-2023-chatgpt}. At the same time, much recent work has indicated that for a broad range of languages, this approach is unfeasible: even SOTA frontier models are simply not capable of translating languages such as Kalamang (out of the box ChrF scores near zero, \citealt{ICLR2024_52d63f9e}) or Owens Valley Paiute \cite{coleman-etal-2026-comparing}, display zero syntactic competence on languages such as Kamtapuri \cite{das-2026-quantifying}, and cannot classify documents into simple topics in dozens of languages \cite{adelani-etal-2024-sib}.

Due to these limitations and little prospect that sufficient amounts of data for in-parameter linguistic knowledge will emerge in such languages, recent studies have experimented with in-context-learning (ICL) approaches which instead provide necessary information for low resource machine translation (LRMT) as part of the prompt. While some papers have successfully enriched LLM prompts with bilingual dictionary information (possible senses of input strings, \citealt{ghazvininejad2023dictionary,lu-etal-2024-chain,zhang-etal-2024-teaching}), it is easy show the limitations of the approach \cite{court-elsner-2024-shortcomings}: individual lexical items are ambiguous (e.g.~the string `lead' in English can mean very different things), effectively forcing models to piece together a plausible sentence using available senses, without the ability to reason over source language grammatical structure. For example, consider the following sentence:

\ex. \textit{the just judges pen firm drafts}\label{ex:sense-sentence}

Although fairly unambiguous, its constituent strings are highly polysemous:

\ex. \{drafts, firm, judges, just, pen, the\}\label{ex:ambig-senses}

The items in \cref{ex:ambig-senses} have many meanings: \textit{just} can be an adverb (just now) or adjective (fair), \textit{judges} can be a noun or verb, \textit{pen} can be a noun or verb, The firm judges just pen drafts. can be an adjective or noun (=company) and \textit{drafts} can be a verb (to recruit) or multiple noun senses (sketches, beer, recruitments, or even teams of freight animals). Asked to create a sentence from \cref{ex:ambig-senses} using each word once, an LLM could easily come up with `The firm judges just pen drafts', rather than the original \cref{ex:sense-sentence}.

As a result of such ambiguities, several of the most recent studies have suggested supplementing possible dictionary entries with some form of grammar information or grammatical analysis \cite{pei-etal-2025-understanding,purushothama-etal-2026-syntax,pei2026reasoninggrammarsyntheticlinguistic}, with approaches ranging from using morphosyntactic glosses (parts of speech or morphological categories), to pasting sections from a grammar book of the source language, to more structured NLP representations, such as syntax trees. However, papers typically target one or two languages with a highly tailored setup, often coupled with gold standard annotations, and there has been no study of multiple languages using a standardized, \textit{predicted} NLP pipeline (e.g.~using generic taggers and parsers in a single formalism).

In this paper we aim to fill this gap, focusing on low resource languages with very low web presence (very small amounts of unlabeled data) but some morphosyntactically labeled data from linguistic annotation projects that can be used to train taggers and parsers (class 2 or `hopeful' languages in the taxonomy of \citealt{joshi-etal-2020-state}). We include five typologically diverse languages: Chukchi (6K words in treebank), Beja (11K), Chintang (14K), Xibe (15K) and Sahidic Coptic (60K). Specifically, we evaluate \textbf{\textsc{UD-PrMPT}} (Universal Dependencies Prompts for Resource-Minimal Parser Translations), a paradigm leveraging morphosyntactic tagging and dependency parsing in the framework of Universal Dependencies (UD, \citealt{de-marneffe-etal-2021-universal}) and test different ICL strategies to include them, comparing the impact of gold standard and predicted NLP parses. We will make the code and publishable lexicon data associated with the paper publicly available.

\footnote{The code and data associated for the paper are available at \url{https://github.com/gucorpling/ud-prmpt/tree/main}.}
Our main contributions are:

\begin{itemize}
    \item A multilingual evaluation of \textsc{UD-PrMPT} 
    \item Comparisons of several serialization strategies
    \item Quantification of the impact of predicted NLP in realistic, very low resource settings 
    \item Release of standardized datasets, processed machine readable lexicons, and splits for MT in five languages
    \item Error analysis across open weight and frontier models 
\end{itemize}

\section{Related work}
Our paper builds on and contrasts recent and growing work in improving ICL-based approaches for low-resource machine translation. Here, we discuss these works and how they relate to this paper.

\paragraph{Dictionary based augmentation} \citet{ghazvininejad2023dictionary} proposed \textsc{DiPMT} (Dictionary-based Prompting for Machine Translation), which selectively lists multiple possible translations for infrequent words based on training data, and evaluates on several medium resource languages from the Flores-101 dataset \cite{goyal-etal-2022-flores}, such as Catalan, Malay, and Norwegian.
\citet{lu-etal-2024-chain} extended this method and proposed Chain-of-Dictionary Prompting, in which multilingual dictionary entries provide word translation options in additional languages. They evaluate on lower resource languages using the Flores-200 dataset \cite{nllbteam2022languageleftbehindscaling}, including Serbian, Sorani Kurdish, and Uyghur. \citet{zhang-etal-2024-teaching} expanded the paradigm to focus on languages with little or no in-parameter representation (out of distribution in training data), and propose \textsc{DiPMT++}, targeting Zhuang and Kalamang using relatively small lexicon data. This paper's use of fuzzy matching to account for inflected and otherwise modified lexical items most closely parallels some of our prefix matching dictionary augmentation methodology.

\paragraph{Grammar based augmentation} Although our paper focuses on use of full syntactic parses of input sentences, multiple forms of grammar based augmentations for ICL have recently been tested. \citet{ICLR2024_52d63f9e}  proposed that adding information from a grammar of Kalamang for ICL prompting could improve translation quality, though subsequent work has suggested improvements may be due to parallel few shot examples, rather than grammar descriptions \cite{ICLR2025_20f44da8}. Similar approaches were tested by \citet{hus-anastasopoulos-2024-back} for a broader range of languages using FLORES-200. 

\paragraph{Morphosyntax and parsing} More recently, papers have tested the idea that morphosyntactic glosses or even full parses could support ICL approaches in LLM based LRMT. In addition to experimenting with grammar prose inputs, \citet{pei-etal-2025-understanding} added morphological analyses for LRMT of Manchu, an agglutinative language, and used LLMs to perform chain-of-thought reasoning, adding morphological and syntactic features to segmented affixes. While the approach is helpful for Manchu, it is highly tailored to the language and cannot easily be extended to other scenarios. 

Only two papers have attempted to use UD trees in LLM-based LRMT. 
In the closest paper to the present work, \citet{purushothama-etal-2026-syntax} directly introduced gold standard or predicted UD parses into translation inputs for Sahidic Coptic in addition to lexicon based augmentation, demonstrating quality improvements over baseline and lexicon-only models. However, their evaluation also used tools tailored to Coptic, such as a syntactic construction module, and they evaluated on only one language. \citet{pei2026reasoninggrammarsyntheticlinguistic} subsequently used UD inputs to approach LRMT for two very low resource languages, Chintang and Xibe, using gold standard UD trees and the UD treebanks' online documentation, as well as $\sim$80 hand-tailored grammar rules per language. Their approach leveraged generating linguistic reasoning traces to analyze sentence syntax in light of the rules and documentation before producing a final translation. Since only gold-standard parses were used in the evaluation, and sentences longer than 30 words were excluded from testing, the resulting performance does not reflect realistic translation accuracy.

The current study expands on these papers by increasing the scope of evaluation to five languages, including Chintang, Coptic, and Xibe to match previous work. It focuses on two important properties: using a completely generic approach to morpho-syntax (just UD inputs, no tailored rules, documentation, or grammar books) and evaluation under realistic settings, using predicted part-of-speech (POS) tags, morphology and UD parses, as well as fuzzy matching following \citet{zhang-etal-2024-teaching}, rather than tailored morphological analyzers.

\section{Languages and Data}

This paper evaluates five typologically diverse languages, drawing on heterogeneous lexicons and treebanks. Below, we describe our criteria for language selection and the sources of the lexicons and treebanks used in our experiments.
\paragraph{Languages} In order to test the potential of morphosyntactic parsing to support LRMT, we focus on low resource languages with very low web presence but labeled data from annotation projects that can be used to train taggers and parsers. We select languages using the following criteria:
(i)~the language must have a treebank published in Universal Dependencies (as of v2.18).
(ii)~the treebank must have sufficient tokens to support rudimentary parser training -- we use at least 2k tokens as a threshold. (iii)~the language must be unfamiliar to existing models. We test this informally by zero-shot prompting several models in chat-bot interfaces such as ChatGPT and Google Gemini, asking to translate a sentence into English.
Unfamiliarity criteria was met if we intuitively judged the model's translation to be substantially inaccurate. 
(iv)~the language must have an accessible dictionary for a lexical component, supporting a reasonable \textsc{DiPMT} baseline. 

The four languages best fitting these criteria were Beja (ISO 639-3: bej), Chuckhi (ckt), Chintang (ctn) and Xibe (sjo). 
In addition to these four languages, we also include Sahidic Coptic (cop) as a reference language, despite chatbots producing partially usable translations for some examples, since it was the language of study in \cite{purushothama-etal-2026-syntax} and otherwise clears all the requirements used above. However, results below indicate that frontier models do possess some, albeit limited, in-parameter knowledge of the language.
 
\paragraph{Lexicon data} For Coptic, \citet{purushothama-etal-2026-syntax} used a high-quality dictionary \citep{feder-etal-2018-linked} integrating entries from multiple sources for a cohesive and extensive resource supporting user-interaction. However, similarly extensive dictionaries are not available for our four languages of study, and a major contribution of this work is formatting diverse resources into publicly available machine readable lexicon data for our study. 

The lexicon data available for these languages range from topic-wise glossaries (lists of animal names, words for people, common verbs) to organized dictionaries extracted from corpus data. Hence, the lexicon data for these languages are diverse in construction, coverage, and details. The overview of these resources is provided in \cref{tab:lexicon-data}. Given the diverse structure of the sources, we use the content structure utilized by the Dictionaria project\footnote{\urldictionaria} for the purpose of adding glossary and dictionary entries into the prompts, which is the original source of the Chintang data. The format facilitates simple bilingual look-up at the word level (source language--English gloss). 

We provide details of the conversion process in \cref{sec:appendix-lexicon-conversion} and scripts will be included in our code release. Below, we provide brief discussion of each lexicon source and the nature of the data.

\textbf{Beja: Fieldwork compilation} We source the lexicon data for Beja from a hitherto unreleased Beja-English-French dictionary under development~\cite{vanhove_hamidahmed_beja_lexique}. We utilize only the bilingual Beja-English information. The dictionary consists of approximately 7.8k entries including clitics and affixes. In the target format, this converts to more than 12k entries to search from.

\textbf{Chukchi: Thematic glossary} For Chukchi, we use the multilingual thematic glossary~\cite{weinstein_chukchi_lexique}. This glossary is part of the Chukchi miscellany collection which includes grammar, literature, and notes of illustrative materials. Unlike a general use dictionary, this glossary organizes Chukchi words and their glosses (with translations and examples) by topics such as animals and foods. The source records are converted 1-to-1 and become a lexicon of more than 11k entries\footnote{We scrape and utilize the openly available web data as the author and maintainer were not reachable. We do not plan to publish or make the data available ourselves due to its unclear license.}.

\textbf{Coptic: Linked Online Dictionary} The linked online dictionary integrates multiple lexicon sources and is accessible through both a REST API and a web interface that visualizes lexical links and frequency information. The dictionary contains searchable definitions in English, French, and German, and provides more than 9k entries in the target format.

\textbf{Chintang: Dictionaria Publication} Perhaps the most organized lexicon data source among the four study languages, the Chintang dictionary~\cite{dictionaria-chintang} is derived from transcribed language data making up the Chintang corpus~\cite{balthazar2004ChintangCorpus}. The dictionary is available in the target format and contains more than 10k entries. 

\textbf{Xibe: Recorded dictionary}
Mini Buleku~\cite{kodner_meng_2021_minibuleku} is a recorded dictionary of Xibe, constructed by its authors based on previous dictionaries, containing multiple forms including transcription of spoken forms, written forms, romanization, and IPA. The converted dictionary contains approximately 4k entries.

\paragraph{UD Treebanks} The four languages were selected alongside Coptic primarily for UD treebank availability, which ensures a consistent augmentation process through a shared annotation framework and label set. However, the treebanks differ in source texts, genres\footnote{\urludgenrepage}, and annotation granularity.

The UD Beja Autogramm treebank~\cite{kahane-etal-2021-morph} was constructed from the IGT glosses of recordings in the language and is hence spoken. The UD Chukchi HSE treebank~\cite{tyers-mishchenkova-2020-dependency} also annotates natural spoken data.
Unlike the other two treebanks, the UD Chintang CTNTB treebank~\cite{tulchynska_etal_2025_ud_chintang} was constructed from grammar examples rather than explicitly written or spoken texts. The UD Xibe XDT treebank~\cite{zhou-etal-2020-universal} consists of texts from multiple sources: grammar examples, newspapers, and language textbooks. The UD Coptic Scriptorium TB data spans a range of written genres including Bible, sermons, letters, and hagiography \cite{zeldes-abrams-2018-coptic}.

\begin{table}[]
\small
    \centering
    \begin{tabular}{l l c}
    Language & Kind  &  Size\\
    \toprule
    Beja & Fieldwork Compilation  & 12k \\
    Chukchi &  Thematic Glossary & 11k\\
    Coptic & Linked Online Dictionary & \hphantom{1}9k\\
    Chintang & Dictionaria Publication & 10k\\
    Xibe & Recorded Dictionary & \hphantom{1}4k\\
    \bottomrule
    \end{tabular}
    \caption{The lexicon data for each language, a kind classification for reference, and the number of first-level entries in the target machine-readable dictionary format.}
    \label{tab:lexicon-data}
    \vspace{-2em}
\end{table}

\section{\textsc{UD-PrMPT}}

In-context learning (ICL) approaches to LRMT aim to compensate for LLMs' limited in-parameter knowledge of the source language by supplying lexical and grammatical information in the prompt, potentially enabling translation into the target language, in this case English, without requiring source-language fluency.

This requires verbalizing potentially ambiguous dictionary entries and generating a grammatical analysis of the input sentence. In this section, we describe our experimental setup, including the augmentation settings and tool pipeline.

\subsection{Experimental setup}
\label{sec:experiment-setup}
\paragraph{Lexicon augmentation}\label{sec:experiment-lexicon} 

Building on \citet{ghazvininejad2023dictionary}, \citet{purushothama-etal-2026-syntax} incorporated multi-sense information where available, as well as custom knowledge from a complex, machine-readable Coptic dictionary. To develop a language-agnostic framework applicable across five typologically diverse languages, we simplify this approach by converting all dictionary data into a common lookup format and using shared code to generate lexical augmentation. The framework still allows for multi-sense matches per word whenever available. We refer to settings involving dictionary augmentation as \lexicon.

We implement an optional fuzzy matching strategy which applies when string lookup fails (no entry matches for a word). We also implement \textbf{prefix} matching to help improve the rate of augmentation for complex, polysynthetic Chukchi.

\paragraph{Syntax augmentation}

Our UD augmentation is similar to but much simpler than \citet{purushothama-etal-2026-syntax}, which utilized a language specific toolkit. Rather than utilizing ~\citeposs{zeldes-schroeder-2016-nlp} implementation of an NLP pipeline for Coptic, we use generic taggers and parsers. We then serialize NLP outputs in three settings that augment \lexicon~with either \dep -- a natural language description of trees and POS tags (``\textit{X (NOUN) is the nominal subject of Y. Y (VERB) is the root.}''), \conll -- the raw 10-column Universal Dependencies tabular format, or both (\syntax). See Appendix \ref{sec:app-prompt} for exact prompts.

\paragraph{Splits} With the exception of Coptic, which has 60K treebanked tokens, the other four languages have fewer than 20k tokens. As 20k is the threshold for providing \texttt{train/dev/test} splits for a UD treebank, these languages only provide a \texttt{test} split. We create custom splits for these, requiring \texttt{test} and \texttt{dev} to each have at least 2k tokens and at most 30\% of the data. The remaining data (no more than 70\%) is used as \texttt{train} for our NLP pipeline, and is not used for LLM experiments. For Coptic, we use the provided three splits for our experiments.

\paragraph{Gold treebanking} In our \texttt{gold} setting, we use the manually annotated `gold' UD treebank of each language for augmentation, specifically including the POS tag, morphological features, and dependency relations of each word, along with the word it depends on. This provides a ceiling for the contribution of morphosyntactic information beyond the more realistic pipeline of having to train taggers and parsers, discussed below.

\subsection{Taggers and parsers} In realistic settings we will not have access to gold treebank data for inputs, so we evaluate the same model setups with predicted POS tags, morphological features, and parses. 

\paragraph{MarMoT tagger}\label{sec:marmot-taggers} Since we have little training data, and our languages are not well supported with existing pretrained embeddings, we employ pre-neural statistical NLP tools. For predicted POS tags and morphological features, we implement separate MarMoT models \cite{mueller2013}\footnote{We use the provided segmentation for training and do not implement a new segmenter.} for each language. The evaluation results for these taggers can be seen in \cref{tab:nlp-test}~\footnote{\label{fn:ud}The eval script is from~\citet{zeman-etal-2018-conll}}. Note that Chukchi does not utilize any morphological features (UFeats) in its treebank. The features and POS related performance around $\simeq85$ is close to the performance of the baseline UDPipe~\cite{straka-etal-2016-udpipe} system from the 2018 shared task on multilingual parsing from raw text~\cite{zeman-etal-2018-conll}. There is room for improvement in the predicted tags and features; errors in predicted tags and features may limit the benefits of prompt augmentation, and contribute to the gaps seen between gold and predicted results. Full details are in \cref{sec:appendix-marmot}.

\paragraph{MaltParser} For parsing, we train lexicalized data-driven dependency parsers using MaltParser~\cite{nivre-etal-2006-maltparser}. We use the \texttt{train} split with gold POS tags, morphological features, and dependency relations for each language to train each parser.  We additionally optimize the parsing algorithm and features for each of the five languages, utilizing the recommendations for optimization provided in the documentation of the tool. We perform this optimization with the \dev~split as the optimization target. Full details are outlined in \cref{sec:appendix-malt}. We evaluate the dependency relations produced using predicted POS tags and morphological features from the MarMoT taggers, hence the evaluation results represent a combined performance. The performance over the \texttt{test} split can be seen in \cref{tab:nlp-test}\footref{fn:ud}. The parsing performance here is close to baseline performance for `small' treebanks from the first version of the reference shared task~\cite{zeman-etal-2018-conll}.

As seen in \cref{tab:nlp-test}, across both tagger and parser, the performance for Coptic is clearly in a class of its own. We attribute this to the size of treebank (with a \texttt{train} split of more than 37k tokens, an order of magnitude more than the largest split from the other four languages).

\begin{table}[t]
\small
\centering
\setlength{\tabcolsep}{4pt}
\begin{tabular}{lrrrrr}
 & bej & ckt & cop & ctn & sjo \\
\midrule
\multicolumn{6}{l}{\textbf{MarMoT}} \\
UFeats & 87.98 & --    & 96.19 & 85.30 & 89.93 \\
UPOS   & 93.28 & 81.86 & 95.92 & 91.42 & 93.13 \\
\midrule
\multicolumn{6}{l}{\textbf{MaltParser}} \\
UAS  & 79.26 & 67.54 & 85.65 & 82.87 & 76.33 \\
LAS  & 65.99 & 43.94 & 77.28 & 57.36 & 69.00 \\
BLEX & 54.32 & 42.46 & 65.96 & 48.22 & 62.82 \\
\bottomrule
\end{tabular}
\caption{Evaluation metrics on \texttt{test} from predicted parses (\% accuracy). UD Chukchi (ckt) does not use UFeats.}
\label{tab:nlp-test}
\vspace{-2em}
\end{table}

\subsection{Models} 
We evaluate one closed-weight frontier model and two open-weight models from families commonly used in prior ICL MT research: GPT, Gemma, and Qwen. From the GPT family, we use the newer GPT-5.6 Terra in place of GPT-4.1. For Gemma, we select Gemma-4-31B, which is comparable in size to the Gemma 3 models evaluated in previous work. For Qwen, we use the slightly newer Qwen3.5-27B. Although larger than previously evaluated Qwen models, it is not explicitly multimodal like its successors.
Respective system cards are referenced in our inference setup in \cref{sec:appendix-model-and-implementation}. 

\subsection{Metrics}
\label{sec:metrics}
We use BERTScore~\cite{bertscore}, specifically the Avg. F1 of BERTScore, as our primary metric for our development work and analyses in this paper. The metric has been shown to have correlation with human evaluation, outperforming many LLM-as-judge settings for MT~\cite{lavie-etal-2025-findings}. We also report BLEU~\cite{papineni-etal-2002-bleu} and ChrF++~\cite{popovic-2017-chrf} as secondary metrics using \texttt{sacrebleu}~implementation~\cite{post-2018-call}\footnote{The \texttt{sacrebleu} signatures for metrics are BLEU:~\path{nrefs:1|case:mixed|eff:no|tok:13a|smooth:exp|version:2.6.0} and ChrF++:~\path{nrefs:1|case:mixed|eff:yes|nc:6|nw:2|space:no|version:2.6.0}}, to supplement the evaluation. These three metrics span character, token, and embedding-based metrics for MT evaluation. The combination can provide us a more complete view of each approach.

\section{Results}

\begin{figure*}[t]
    \centering
    \includegraphics[width=\textwidth]{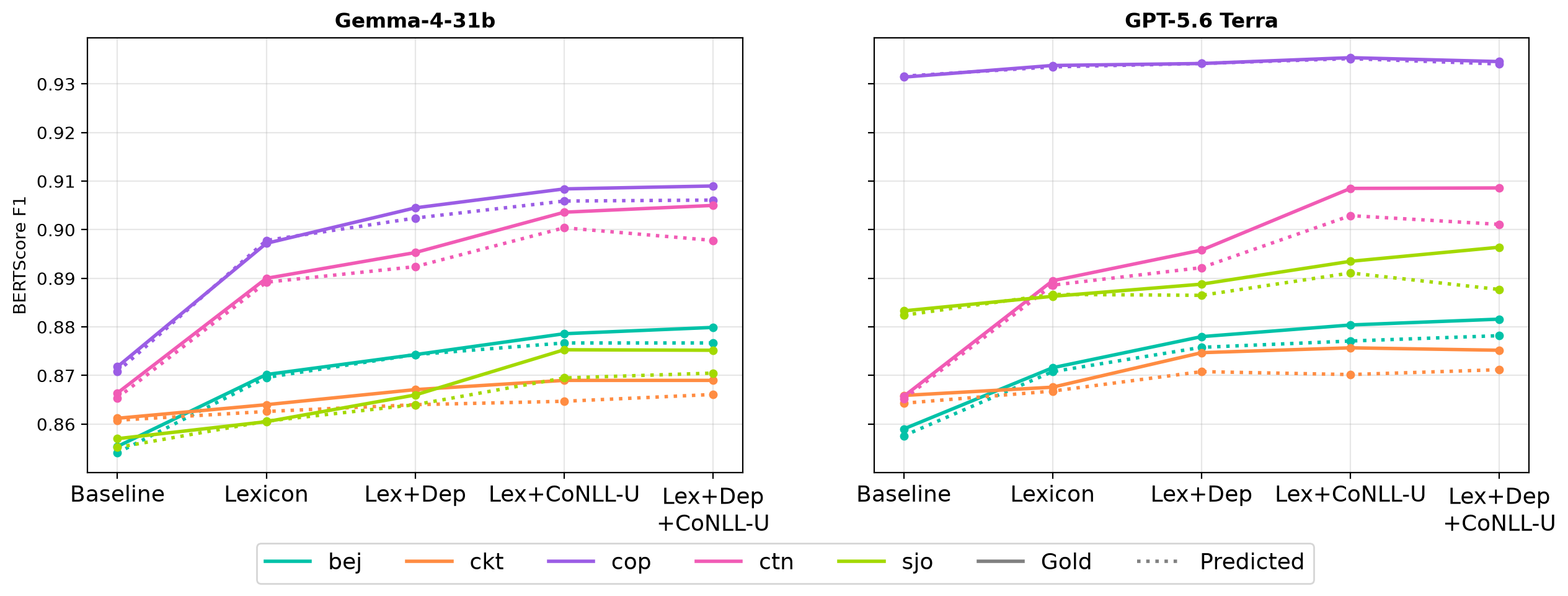}
    \caption{BERTScore F1 for translations generated using gold versus predicted linguistic annotations across two language models and five prompting settings. Results for Qwen are omitted for visual clarity because its trends closely resemble those of Gemma.}
    \label{fig:gold_pred_results}
    \vspace{-1em}
\end{figure*}

For all languages and models, \textbf{\textsc{UD-PrMPT}} significantly improves translation quality over the baseline. As expected, adding \lexicon~provides the largest single gain, but further improvements are possible with the addition of syntactic information. At least one configuration incorporating \dep, \conll, or both significantly outperforms \lexicon~alone for every language and model, even when using predicted rather than gold NLP outputs. These trends can be observed in Figure \ref{fig:gold_pred_results}, with full tables reported in Appendix \ref{sec:app-gold-results}.

Among syntactic configurations, \lexicon +\conll~ generally outperforms \lexicon+\dep, and the best-performing configuration for each language+model pair is either \lexicon+\conll~or \lexicon+\dep+\conll. The benefit of combining both syntactic components, however, is inconsistent. In the significance analysis, \lexicon+\conll~outperforms \lexicon~in every language/model pair except Coptic with GPT, and likewise for \lexicon+\syntax~ (paired $t$-tests with Holm correction; Appendix~\ref{sec:appendix-sig}). By contrast, differences between syntactic configurations are not consistently significant. We hypothesize that raw CoNLL-U tables outperform natural language dependency descriptions (DEP) because dense, structured column layouts provide consistent structural anchors and clear token-index pointers for attention mechanisms, though the choice of syntactic augmentation may matter less than its inclusion at all.

\paragraph{Languages} While the trends hold per language, it is easy to see that some languages fare better than others. In particular, Coptic appears to have the highest overall scores, particularly when using the Gemma and even more so the GPT model, which appears to have some in-parameter command of the language. After Coptic, Chintang sees the biggest gains from the addition of each setting, which is striking given what appears to be qualitatively near-zero knowledge of the language. Beja shows consistent but smaller gains. Chukchi and Xibe on the other hand do not see strong gains from \lexicon, though Xibe is aided by syntactic components where Chukchi is not as influenced. This can be partly attributed to the difficulty in lexicon augmentation that we observe during development, and mitigated to a limited extent with prefix matching. 

For Chukchi and Xibe we see low `hit' rates of lexicon augmentation, $<10\%$ at token level (augmentation happens for one of 10 tokens in source) and $<17\%$ and $<48\%$ at sentence level respectively (proportion of \dev~split sentences for which at least one definition is included.). For comparison, the rates for other three languages are $>60\%$ at token level and $>95\%$ at sentence level.  This is partly responsible for the two languages seeing the least gains by \lexicon~only. In \textsc{UD-PrMPT}, this indicates two challenges for Chukchi, one of morphology and another of lexicon. Chukchi exhibits complex morphology and is considered polysynthetic~\cite{tyers-mishchenkova-2020-dependency}, given that high coverage in lexical augmentation becomes a challenge in the absence of high quality morphological segmentation. The second is a question of `domain' and resource diversity. The Chukchi dictionary we use is explicitly a thematic glossary which has many mismatches to the treebank made from transcribed conversations.

\paragraph{Models} Although more expensive and less reproducible, GPT-5.6 Terra achieves the best overall performance, followed by Gemma-4-31B and Qwen3.5-27B. Performance is relatively consistent across models for Beja, Chintang, and Chukchi, whereas model choice has a greater impact on Coptic and Xibe. For Coptic, GPT and Gemma significantly outperform Qwen, consistent with the possibility that these models have greater prior exposure to the language during training. This interpretation is also supported by GPT's relatively strong baseline performance on Coptic compared with the other languages.

\paragraph{Predicted NLP: Even baseline tools provide improvements}
As outlined in \cref{sec:experiment-setup}, we also evaluated our models using predicted POS, morphological features and dependency parses. Despite the trained taggers and parsers showing intrinsic performance at baseline level due to the small amounts of training data, they still provide gains in all languages. Even when using baseline-equivalent tools in the predicted NLP settings, the addition of any syntactic component on top of \lexicon~provides gains.
 
As Figure \ref{fig:gold_pred_results} shows, the difference between gold and predicted NLP features is very small for some languages (notably Beja, and Coptic except for Qwen), but more subtantial in other cases (especially Xibe, and Chintang except for Qwen). The smaller deltas for Qwen, our smallest model, suggest it may be less able to exploit the morphosyntactic information fully. 
Full numerical results are provided be found in Appendix \ref{sec:app-pred-results}.
\begin{table*}[h!tbp]
\footnotesize
\centering
\setlength{\tabcolsep}{3pt}
\begin{tabular}{@{}p{2.3cm}p{4.6cm}p{4.6cm}p{4.6cm}@{}}
\toprule
\textbf{Lang+Model} & \textbf{Reference} & \textbf{\lexicon} & \textbf{\all} \\
\midrule
\textbf{Chintang} Gemma & I live up there, on top of the ravine. & That is where the ravine is. & Oh, I live up there in the ravine. \\
\textbf{Chintang} Qwen & Kalpana pulls out people's hair! & Pull out Kalpana's hair! & Kalpana is pulling out my hair! \\
\textbf{Chintang} GPT & It's said they didn't speak for a while. & Please speak for a moment. & They reportedly did not speak for a while. \\
\midrule
\textbf{Beja} Gemma & They wanted to eat him. & Those are the big ones. & They wanted to kill him. \\
\textbf{Coptic} Gemma & They are like the light of day. & And they likened it to the light of the day. & And they were like the light of the day. \\
\textbf{Xibe} Gemma & Let us go sightseeing to the Ili River! & The yellow flowers bloom in the spring, and the wind blows gently! & Let us go to the valley of the Elbi river! \\
\textbf{Chukchi} Gemma & That is \guillemotleft Mrenti\guillemotright. & The child is sleeping. & It's Mrenti. \\
\bottomrule
\end{tabular}
\caption{Selected \texttt{dev} outputs under \lexicon\ and \all. Chintang is the focus language. Full examples including baseline outputs and $\Delta$F1, are in Appendix~\ref{sec:app-examples}.}
\label{tab:err-ana-sample}
\vspace{-1em}
\end{table*}

\paragraph{Better NLP, better \textsc{UD-PrMPT}} 
We also note that parsing accuracy varies substantially across the five languages as outlined in Table \ref{tab:nlp-test}. We observe that languages with worse parser performance (such as Chukchi) show weaker gains when using predicted data over gold. Morphological features were not available in the training data for Chukchi, which might have been helpful to models as well.

Although predicted NLP settings never beat gold, they demonstrate that syntactic augmentation can show strong results downstream even without perfect annotations. Our parsers were not tuned exhaustively, so while the gap between gold and predicted results is likely an upper bound, treebank and parser development would help narrow it. Even so, predicted annotations recover much of the gains seen in  gold settings, and the remaining gap, which is largest where parsing is weakest, motivates continued investment in gold-standard resources for these languages. 

\section{Analysis and examples}

We analyze gold \texttt{dev} translations to qualitatively examine gains and errors, comparing each output to the English
reference. Chintang is our focus language because it showed the largest \lexicon$\rightarrow$\all\ gain in \texttt{dev} experiments. We then check whether the same patterns hold in the other four languages. Table~\ref{tab:err-ana-sample} shows selected outputs; the full examples with $\Delta$F1 are in Appendix \ref{sec:app-examples}.

\paragraph{Chintang}
The most consistent effect of syntax is recovering the subject and its role. With \lexicon\ alone, models often drop or misassign the subject. Gemma's \lexicon\ output (``That is where the ravine is'') loses the
first-person subject, which \all\ restores (``I live up there''). For the Kalpana sentence, Qwen's \lexicon\ output makes Kalpana the
possessor or patient (``Pull out Kalpana's hair''), whereas \all\
correctly makes her the agent, although the possessor is still wrong
(``my hair'' instead of ``people's hair''). The same holds for argument
structure: only with syntax does Gemma produce the three-participant event ``Anita introduced Lokendra to Lakhman,'' although the reciprocal (``to each other'') is lost (Appendix~\ref{sec:app-examples}). With GPT, morphosyntactic information also recovers negation, person and number, and evidentiality (``reportedly'') that \lexicon\ alone misses or mistranslates.

\paragraph{Other languages}
Using Gemma, the same patterns appear in the other languages. In Beja,
\lexicon\ yields an unrelated sentence, whereas \all\ recovers the
subject, object, and the ``want to'' frame, although the verb is wrong
(``kill'' for ``eat''). In Coptic, \all\ corrects the agent/patient
analysis of a comparison (``they likened it to the light of day''
$\rightarrow$ ``they were like the light of day''). In
Xibe, \lexicon\ again yields an unrelated sentence, while \all\ recovers ``Let us go to\ldots'' and the destination, though the place name is approximated (``Elbi river'' for ``Ili River'') and ``sightseeing'' is lost. In Chukchi, \lexicon\ yields an unrelated sentence (``The child is sleeping''), while \all\ recovers the copular clause and the name (``It's Mrenti''), though notably this sentence is very short and contains a proper name.

\section{Conclusion}

We present \textsc{UD-PrMPT}, a method of augmenting lexicon-based approaches with syntax derived from Universal Dependencies, to improve machine translation from five low-resource languages to English. We evaluate five typologically diverse languages (Beja, Chintang, Chukchi, Coptic, and Xibe) using both gold data and predicted data obtained from generic taggers and parsers. We find that adding lexicon inputs significantly improves translations over a zero-shot baseline, and crucially observe that the addition of a morphosyntactic component also significantly improves translation quality over pure lexical augmentation. Qualitative examples from our analysis show that the addition of syntax helps restore certain grammatical structures from the source sentence that are not recovered from lexicon alone.

We find that using predicted parses provides comparable results, suggesting that generic taggers and parsers based on frameworks such MarMoT and Malt are sufficient for this task, though we still see a gap to the gold treebank ceiling. We release all code including tagger and parser training and models to the community for research and practical work on LRMT in these and other languages.

\section*{Limitations}

\paragraph{Lack of native speakers}
Our team does not include native speakers of any of the five source languages examined in this study, which limits our ability to critically evaluate prompts and translation outputs. We therefore conducted our analysis on the English translations, a language in which all authors are fluent, and supported it with the linguistic annotations of source-sentence structure provided by the UD source treebanks.

\paragraph{Lexicon quality}
For Chintang and Coptic, we used curated lexicon data. For the other three languages, we derived lexicons from web sources and converted them into our format using deterministic, automatic rules developed through AI-assisted exploration of the source data, followed by limited manual verification. Consequently, the lexicons for these three languages are of silver rather than gold quality, and errors or gaps in them may affect downstream translation results.

\paragraph{Data contamination}
We did not perform a contamination analysis, and it is possible that the models we evaluated were exposed to these datasets during training. This is especially likely for GPT and Coptic, where the model appears to have substantial knowledge of the language and may have seen the specific texts or treebanks used here.

\paragraph{Inference variance}
We observed higher variance in outputs when querying models through OpenRouter than when running local implementations. We have not conducted additional experiments to quantify this difference.

\paragraph{Dataset size and evaluation}
As in our prior work, the datasets we use are small, which limits the statistical power and generalizability of our findings. In addition, the metrics we rely on have not been validated for translation from low-resource languages into English, and we did not conduct human evaluation of the translations, relying solely on automatic metrics.

\section*{Acknowledgments}
Compute and inference resources provided through the Georgetown University Department of Linguistics were used for this paper. We would like to thank the developers of the dictionaries used in this paper for making their lexical resources available online.  

\bibliography{custom}

\appendix
\section{Additional details}

\paragraph{Lexicon data conversion}
\label{sec:appendix-lexicon-conversion}
Dictionaria uses a three file dictionary structure; for our use, only two of the tables (namely `entries' and `senses') are required. We explore each of the sources and convert their data into this format. For Beja, following the examples of the Chintang data, we converted the SIL MDF format data~\cite{coward_grimes_2000_mdf} into the Dictionaria structure. For Chukchi the data was extracted based on the HTML structure used in the source website, and then converted the result to the final format. For Xibe, the same was done, but from embedded JSON data from the lexicon's website. All three conversions utilized AI coding assistance to different degrees. However, for replicability, the logic was written as a single script and is part of the published code on our repository. 

\subsection{Prompt sample}
\label{sec:app-prompt}
\begin{figure*}[tbp]
    High-level instruction, adapted from \citet{purushothama-etal-2026-syntax}.
    \begin{tcolorbox}
        You are a professional Chintang-to-English translator tasked with providing translations suitable for use in United States (en\_US). Your goal is to accurately convey the meaning and nuances of the original Chintang text while adhering to English grammar, vocabulary, and cultural sensitivities.Produce only the English translation, without any additional explanations or commentary.Please translate the following Chintang text into English (en\_US):\textbackslash{}nEkchin macekuyoktacepho.

\textbackslash{}nUsing all the information provided above, now please translate the sentence into English(en\_US). Remember your source sentence is: Ekchin macekuyoktacepho. The English translation is:
    \end{tcolorbox}
    \lexicon
    \begin{tcolorbox}
        For the translation task, you are given dictionary entries for Chintang. Some words can be polysemous and there might be multiple entries. Note that for some words, they might be derived from a more basic form, some entries will be for such lemma. Here are the entries for collected for individual words in the sentence: \\ Dictionary entry adv ekchin has 1 senses.\\ Sense 1:\\ Ekchin is glossed as one.moment; little.while\\ Dictionary entry vt cekma has 1 senses.\\ Sense 1:\\ macekuyoktace is glossed as say; speak; accept; believe.in\\
    \end{tcolorbox}
    \dep
    \begin{tcolorbox}
        The dependency information for the sentence is:
        Ekchin (ADV) is the adverbial modifier of macekuyoktace. macekuyoktace(VERB) is the root. pho (AUX) is the auxiliary of macekuyoktace.
    \end{tcolorbox}
    \conll
    \begin{tcolorbox}
    The raw conllu data for the sentence is in the CONLL-U format:\hphantom{abcdefghijklmnopqrstuvwxyz}\\ 
    \begin{tabular}{@{}llllllllll@{}}
1   & Ekchin           & ekchin & ADV   & \_  & \_ & 2 & advmod & \_ & \_ \\
2-3 & macekuyoktacepho & \_     & \_    & \_  & \_ & \_ & \_    & \_ & \_ \\
2   & macekuyoktace    & cekt   & VERB  & \_  & Aspect=Perfv\ldots & 0 & root & \_ & \_ \\
3   & pho              & pho    & AUX   & REP & Evident=Nfh & 2 & aux   & \_ & \_ \\
4   & .                & .      & PUNCT & \_  & \_ & 2 & punct  & \_ & \_ \\
\end{tabular}
    \end{tcolorbox}
    \caption{A sample of prompt elements for Chintang (ctn) with \lexicon~, \dep~, and \conll~ using predicted syntax trees.}
    \label{fig:prompt-text}
\end{figure*}

\subsection{Models and implementation}
\label{sec:appendix-model-and-implementation}

The specific card or report for models in our experiments are in \cref{tab:models}.

\paragraph{OpenRouter platform} We performed all our experiments using the OpenRouter platform\footnote{\urlopenrouter.}. The platform allows for inference for both open and closed models, across many inference providers. For the two open models, we configured parameters to limit variance such as fixing the random seed. For GPT-5.6-Terra, such configuration was not supported.

\subsection{MarMoT}
\label{sec:appendix-marmot}

For each language, the \texttt{train}, \texttt{dev}, and \texttt{test} CoNLL-U files are first converted to the \texttt{tab} format required by the MarMot parser. Then, a MarMot model is trained on the converted materials. The model is then used to predict POS tags and morphological feats on \texttt{dev} and \texttt{test}. The final predicted files are then converted back from the \texttt{tab} format to CoNLL-U for use in the MaltParser pipeline, described in Appendix \ref{sec:appendix-malt}. 

\begin{table}[t]
    \centering
    \begin{tabular}{ll}
         \textbf{Model} & \textbf{System Card or Report}\\
         \toprule
         GPT-5.6 Terra &  \href{https://deploymentsafety.openai.com/gpt-5-6}{OpenAI System Card}\\
         Gemma-4-31B (-it) & \citet{gemmateam2026gemma4technicalreport}\\
         Qwen3.5-27B & \href{https://huggingface.co/Qwen/Qwen3.5-27B}{Huggingface System Card}\\
         \bottomrule
    \end{tabular}
    \caption{Models used for our experiments. }
    \label{tab:models}
\end{table}

\subsection{MaltParser}
\label{sec:appendix-malt}
For each language, the model is trained on \texttt{train}, optimized on \texttt{dev}, and ultimately evaluated on \texttt{test}. The \texttt{train} set contains gold POS and morphological features, while \texttt{dev} and \texttt{test} utilize predicted features from MarMoT. 

To optimize a Malt model, the creators recommend a series of steps to get the best model. First, the parsing algorithms \texttt{nivreeager} and \texttt{stackproj} are compared. If the language at hand has non-projectivities, it is recommended to try these models with different kinds of pseudo-projective pre- and post-processing.

The learning algorithm can either be \texttt{liblinear} or \texttt{libsvm}; \texttt{liblinear} is recommended when data is small, so this is the algorithm selected for our languages as data is scarce. 

Since data is scarce, we also opt to not split the training data; this aligns with the recommendations outlined in the optimization recommendations. 

Finally, the authors recommend adding explicit features to train the models. Multiple feature sets were tried for each language, with the best result on the \texttt{dev} set being used to train the final Malt model for each language. 

The evaluation script used for MarMoT and Malt is from \citep{zeman-etal-2018-conll}, and is published at \urlconllueval.

\subsection{Full results}
Here we report the complete metrics that inform the results in the main text. For each language (bej, ckt, cop, ctn, sjo), we report BLEU, chrF++, and BERTScore F1 for all three models (Gemma, GPT, and Qwen) under five prompting
settings: the \texttt{baseline} prompt, \texttt{lexicon} alone, and the lexicon
combined with CoNLL-U annotations (\texttt{lexicon-conllu}), dependency
information (\texttt{lexicon-dependency}), or both (\texttt{lexicon-conllu-dependency}). We first report gold and predicted results on the \texttt{dev} set, followed by \texttt{test}.

\paragraph{Dev: Gold}\label{sec:app-dev-gold-results} Tables \ref{tab:dev-bej-gold} through \ref{tab:dev-sjo-gold} provide the full metrics on the \texttt{dev} set with gold annotations.

\begin{table*}[htbp]
\centering
\resizebox{\textwidth}{!}{
\begin{tabular}{llrrr}
\toprule
Setting & Model & BLEU & ChrF++ & BERTScore F1 \\
\midrule
Baseline & Gemma-4-31b & 0.3406 & 14.5752 & 0.8553 \\
Baseline & GPT-5.6-terra & 0.3705 & 14.0662 & 0.8584 \\
Baseline & Qwen3.5-27B & 0.1342 & 12.1829 & 0.8468 \\
Lexicon & Gemma-4-31b & 2.0692 & 21.2349 & 0.8705 \\
Lexicon & GPT-5.6-terra & 1.1704 & 20.4120 & 0.8732 \\
Lexicon & Qwen3.5-27B & 1.5215 & 20.6745 & 0.8645 \\
Lexicon + CoNLL-U & Gemma-4-31b & 3.8482 & 25.0895 & 0.8800 \\
Lexicon + CoNLL-U & GPT-5.6-terra & 4.9469 & 25.3976 & 0.8832 \\
Lexicon + CoNLL-U & Qwen3.5-27B & 3.0857 & 24.1747 & 0.8755 \\
Lexicon + Dependency & Gemma-4-31b & 2.5027 & 22.2634 & 0.8758 \\
Lexicon + Dependency & GPT-5.6-terra & 1.9518 & 20.8843 & 0.8765 \\
Lexicon + Dependency & Qwen3.5-27B & 2.0244 & 20.6350 & 0.8714 \\
Lexicon + CoNLL-U + Dependency & Gemma-4-31b & 4.2750 & 25.5442 & 0.8814 \\
Lexicon + CoNLL-U + Dependency & GPT-5.6-terra & 4.1213 & 24.5618 & 0.8833 \\
Lexicon + CoNLL-U + Dependency & Qwen3.5-27B & 3.3096 & 24.2963 & 0.8780 \\
\bottomrule
\end{tabular}
}
\caption{Beja (bej) \dev results using the gold treebank.}
\label{tab:dev-bej-gold}
\end{table*}

\begin{table*}[tbp]
\centering
\resizebox{\textwidth}{!}{%
\begin{tabular}{llrrr}
\toprule
Setting & Model & BLEU & ChrF++ & BERTScore F1 \\
\midrule
Baseline & Gemma-4-31b & 0.6002 & 14.3879 & 0.8622 \\
Baseline & GPT-5.6-terra & 1.6016 & 13.8054 & 0.8665 \\
Baseline & Qwen3.5-27B & 0.6902 & 12.7648 & 0.8556 \\
Lexicon & Gemma-4-31b & 0.6907 & 15.2981 & 0.8639 \\
Lexicon & GPT-5.6-terra & 1.7808 & 14.8351 & 0.8681 \\
Lexicon & Qwen3.5-27B & 0.7879 & 13.4402 & 0.8589 \\
Lexicon + CoNLL-U & Gemma-4-31b & 1.0307 & 15.6181 & 0.8691 \\
Lexicon + CoNLL-U & GPT-5.6-terra & 2.5863 & 16.6060 & 0.8750 \\
Lexicon + CoNLL-U & Qwen3.5-27B & 1.1438 & 14.8408 & 0.8666 \\
Lexicon + Dependency & Gemma-4-31b & 1.3668 & 15.9755 & 0.8685 \\
Lexicon + Dependency & GPT-5.6-terra & 2.4462 & 16.4271 & 0.8743 \\
Lexicon + Dependency & Qwen3.5-27B & 1.9704 & 15.6643 & 0.8669 \\
Lexicon + CoNLL-U + Dependency & Gemma-4-31b & 1.3603 & 15.9914 & 0.8709 \\
Lexicon + CoNLL-U + Dependency & GPT-5.6-terra & 2.4309 & 16.8506 & 0.8749 \\
Lexicon + CoNLL-U + Dependency & Qwen3.5-27B & 1.6062 & 15.6218 & 0.8680 \\
\bottomrule
\end{tabular}%
}
\caption{Chukchi (ckt) \dev results using the gold treebank.}
\label{tab:dev-ckt-gold}
\end{table*}

\begin{table*}[tbp]
\centering
\resizebox{\textwidth}{!}{%
\begin{tabular}{llrrr}
\toprule
Setting & Model & BLEU & ChrF++ & BERTScore F1 \\
\midrule
Baseline & Gemma-4-31b & 6.6394 & 26.4302 & 0.8713 \\
Baseline & GPT-5.6-terra & 28.9172 & 49.6033 & 0.9286 \\
Baseline & Qwen3.5-27B & 1.9830 & 19.4095 & 0.8493 \\
Lexicon & Gemma-4-31b & 12.6869 & 36.1852 & 0.8981 \\
Lexicon & GPT-5.6-terra & 28.8503 & 50.3020 & 0.9317 \\
Lexicon & Qwen3.5-27B & 10.9625 & 33.2071 & 0.8889 \\
Lexicon + CoNLL-U & Gemma-4-31b & 15.0247 & 40.2140 & 0.9075 \\
Lexicon + CoNLL-U & GPT-5.6-terra & 29.9167 & 51.1698 & 0.9335 \\
Lexicon + CoNLL-U & Qwen3.5-27B & 15.1774 & 39.1918 & 0.9025 \\
Lexicon + Dependency & Gemma-4-31b & 14.4430 & 38.5775 & 0.9049 \\
Lexicon + Dependency & GPT-5.6-terra & 29.5904 & 50.8342 & 0.9326 \\
Lexicon + Dependency & Qwen3.5-27B & 12.3311 & 35.2901 & 0.8961 \\
Lexicon + CoNLL-U + Dependency & Gemma-4-31b & 16.0798 & 40.6677 & 0.9060 \\
Lexicon + CoNLL-U + Dependency & GPT-5.6-terra & 29.7833 & 51.0364 & 0.9334 \\
Lexicon + CoNLL-U + Dependency & Qwen3.5-27B & 16.0097 & 39.9589 & 0.9037 \\
\bottomrule
\end{tabular}%
}
\caption{Coptic (cop) \dev results using the gold treebank.}
\label{tab:dev-cop-gold}
\end{table*}

\begin{table*}[tbp]
\centering
\resizebox{\textwidth}{!}{%
\begin{tabular}{llrrr}
\toprule
Setting & Model & BLEU & ChrF++ & BERTScore F1 \\
\midrule
Baseline & Gemma-4-31b & 0.6085 & 12.8178 & 0.8639 \\
Baseline & GPT-5.6-terra & 1.3617 & 12.7415 & 0.8627 \\
Baseline & Qwen3.5-27B & 0.9082 & 12.3902 & 0.8612 \\
Lexicon & Gemma-4-31b & 4.6436 & 21.9314 & 0.8872 \\
Lexicon & GPT-5.6-terra & 4.3630 & 21.8108 & 0.8851 \\
Lexicon & Qwen3.5-27B & 3.4893 & 20.8321 & 0.8808 \\
Lexicon + CoNLL-U & Gemma-4-31b & 8.5852 & 28.0799 & 0.9023 \\
Lexicon + CoNLL-U & GPT-5.6-terra & 9.8719 & 29.5369 & 0.9049 \\
Lexicon + CoNLL-U & Qwen3.5-27B & 6.6037 & 26.0852 & 0.8973 \\
Lexicon + Dependency & Gemma-4-31b & 5.2974 & 23.3832 & 0.8917 \\
Lexicon + Dependency & GPT-5.6-terra & 5.6345 & 23.6935 & 0.8917 \\
Lexicon + Dependency & Qwen3.5-27B & 4.7136 & 22.5042 & 0.8864 \\
Lexicon + CoNLL-U + Dependency & Gemma-4-31b & 8.5321 & 28.3467 & 0.9026 \\
Lexicon + CoNLL-U + Dependency & GPT-5.6-terra & 9.7385 & 29.6506 & 0.9055 \\
Lexicon + CoNLL-U + Dependency & Qwen3.5-27B & 7.2054 & 26.7276 & 0.8973 \\
\bottomrule
\end{tabular}%
}
\caption{Chintang (ctn) \dev results using the gold treebank.}
\label{tab:dev-ctn-gold}
\end{table*}

\begin{table*}[tbp]
\centering
\resizebox{\textwidth}{!}{%
\begin{tabular}{llrrr}
\toprule
Setting & Model & BLEU & ChrF++ & BERTScore F1 \\
\midrule
Baseline & Gemma-4-31b & 1.2094 & 19.4997 & 0.8525 \\
Baseline & GPT-5.6-terra & 5.2956 & 26.1174 & 0.8802 \\
Baseline & Qwen3.5-27B & 1.2042 & 19.4464 & 0.8455 \\
Lexicon & Gemma-4-31b & 1.0802 & 20.2488 & 0.8567 \\
Lexicon & GPT-5.6-terra & 5.8598 & 27.4408 & 0.8818 \\
Lexicon & Qwen3.5-27B & 1.2875 & 19.8759 & 0.8502 \\
Lexicon + CoNLL-U & Gemma-4-31b & 2.6665 & 23.0625 & 0.8697 \\
Lexicon + CoNLL-U & GPT-5.6-terra & 6.9223 & 29.8831 & 0.8887 \\
Lexicon + CoNLL-U & Qwen3.5-27B & 2.1846 & 21.9844 & 0.8640 \\
Lexicon + Dependency & Gemma-4-31b & 1.5440 & 20.4131 & 0.8615 \\
Lexicon + Dependency & GPT-5.6-terra & 5.4458 & 27.7510 & 0.8849 \\
Lexicon + Dependency & Qwen3.5-27B & 2.7620 & 21.9595 & 0.8609 \\
Lexicon + CoNLL-U + Dependency & Gemma-4-31b & 2.6481 & 23.0456 & 0.8698 \\
Lexicon + CoNLL-U + Dependency & GPT-5.6-terra & 7.6791 & 30.7969 & 0.8907 \\
Lexicon + CoNLL-U + Dependency & Qwen3.5-27B & 2.4383 & 23.1145 & 0.8658 \\
\bottomrule
\end{tabular}%
}
\caption{Xibe (sjo) \dev results using the gold treebank.}
\label{tab:dev-sjo-gold}
\end{table*}

\paragraph{Dev: Predicted NLP}
\label{sec:app-dev-pred-results}
Full metrics on the \texttt{dev} set using gold data can be found in Tables \ref{tab:dev-bej-predicted}, \ref{tab:dev-ckt-predicted}, \ref{tab:dev-cop-predicted}, \ref{tab:dev-ctn-predicted}, and \ref{tab:dev-sjo-predicted}.

\begin{table*}[tbp]
\centering
\resizebox{\textwidth}{!}{%
\begin{tabular}{llrrr}
\toprule
Setting & Model & BLEU & ChrF++ & BERTScore F1 \\
\midrule
Baseline & Gemma-4-31b & 0.3469 & 14.4410 & 0.8555 \\
Baseline & GPT-5.6-terra & 0.3347 & 14.1161 & 0.8588 \\
Baseline & Qwen3.5-27B & 0.2469 & 13.9814 & 0.8490 \\
Lexicon & Gemma-4-31b & 2.2952 & 21.2402 & 0.8704 \\
Lexicon & GPT-5.6-terra & 1.5764 & 20.7902 & 0.8723 \\
Lexicon & Qwen3.5-27B & 1.7327 & 20.6883 & 0.8655 \\
Lexicon + CoNLL-U & Gemma-4-31b & 3.1787 & 24.7194 & 0.8761 \\
Lexicon + CoNLL-U & GPT-5.6-terra & 3.7391 & 24.3437 & 0.8783 \\
Lexicon + CoNLL-U & Qwen3.5-27B & 2.7971 & 23.5727 & 0.8723 \\
Lexicon + Dependency & Gemma-4-31b & 1.9912 & 21.8780 & 0.8734 \\
Lexicon + Dependency & GPT-5.6-terra & 2.5479 & 20.8268 & 0.8746 \\
Lexicon + Dependency & Qwen3.5-27B & 1.8407 & 20.5537 & 0.8697 \\
Lexicon + CoNLL-U + Dependency & Gemma-4-31b & 3.5073 & 24.8513 & 0.8776 \\
Lexicon + CoNLL-U + Dependency & GPT-5.6-terra & 3.3058 & 23.8029 & 0.8784 \\
Lexicon + CoNLL-U + Dependency & Qwen3.5-27B & 3.2257 & 23.3258 & 0.8733 \\
\bottomrule
\end{tabular}%
}
\caption{Beja (bej) \dev results using predicted NLP annotations.}
\label{tab:dev-bej-predicted}
\end{table*}

\begin{table*}[tbp]
\centering
\resizebox{\textwidth}{!}{%
\begin{tabular}{llrrr}
\toprule
Setting & Model & BLEU & ChrF++ & BERTScore F1 \\
\midrule
Baseline & Gemma-4-31b & 0.5015 & 14.3073 & 0.8612 \\
Baseline & GPT-5.6-terra & 1.4337 & 14.0032 & 0.8652 \\
Baseline & Qwen3.5-27B & 0.6922 & 12.8250 & 0.8556 \\
Lexicon & Gemma-4-31b & 0.6723 & 15.1761 & 0.8640 \\
Lexicon & GPT-5.6-terra & 1.8557 & 15.0947 & 0.8677 \\
Lexicon & Qwen3.5-27B & 1.0494 & 14.1055 & 0.8591 \\
Lexicon + CoNLL-U & Gemma-4-31b & 0.8326 & 14.7183 & 0.8646 \\
Lexicon + CoNLL-U & GPT-5.6-terra & 2.3199 & 15.5034 & 0.8695 \\
Lexicon + CoNLL-U & Qwen3.5-27B & 1.4016 & 14.7856 & 0.8632 \\
Lexicon + Dependency & Gemma-4-31b & 1.5154 & 15.2780 & 0.8651 \\
Lexicon + Dependency & GPT-5.6-terra & 1.5470 & 15.3145 & 0.8691 \\
Lexicon + Dependency & Qwen3.5-27B & 1.5482 & 14.9309 & 0.8644 \\
Lexicon + CoNLL-U + Dependency & Gemma-4-31b & 1.0317 & 14.8557 & 0.8661 \\
Lexicon + CoNLL-U + Dependency & GPT-5.6-terra & 1.6207 & 15.0112 & 0.8682 \\
Lexicon + CoNLL-U + Dependency & Qwen3.5-27B & 1.4602 & 14.7011 & 0.8636 \\
\bottomrule
\end{tabular}%
}
\caption{Chukchi (ckt) \dev results using predicted NLP annotations.}
\label{tab:dev-ckt-predicted}
\end{table*}

\begin{table*}[tbp]
\centering
\resizebox{\textwidth}{!}{%
\begin{tabular}{llrrr}
\toprule
Setting & Model & BLEU & ChrF++ & BERTScore F1 \\
\midrule
Baseline & Gemma-4-31b & 6.5727 & 26.3763 & 0.8706 \\
Baseline & GPT-5.6-terra & 28.8392 & 49.6635 & 0.9284 \\
Baseline & Qwen3.5-27B & 2.1839 & 19.9679 & 0.8496 \\
Lexicon & Gemma-4-31b & 13.3913 & 37.0209 & 0.8990 \\
Lexicon & GPT-5.6-terra & 28.9393 & 50.5585 & 0.9321 \\
Lexicon & Qwen3.5-27B & 10.5821 & 33.3743 & 0.8885 \\
Lexicon + CoNLL-U & Gemma-4-31b & 15.0253 & 39.9466 & 0.9058 \\
Lexicon + CoNLL-U & GPT-5.6-terra & 29.7741 & 51.1471 & 0.9332 \\
Lexicon + CoNLL-U & Qwen3.5-27B & 14.2297 & 39.0818 & 0.9002 \\
Lexicon + Dependency & Gemma-4-31b & 14.1551 & 38.5081 & 0.9024 \\
Lexicon + Dependency & GPT-5.6-terra & 29.1711 & 50.6061 & 0.9315 \\
Lexicon + Dependency & Qwen3.5-27B & 12.7367 & 35.5615 & 0.8947 \\
Lexicon + CoNLL-U + Dependency & Gemma-4-31b & 14.6997 & 39.9636 & 0.9051 \\
Lexicon + CoNLL-U + Dependency & GPT-5.6-terra & 29.8909 & 51.0240 & 0.9325 \\
Lexicon + CoNLL-U + Dependency & Qwen3.5-27B & 15.0001 & 39.2746 & 0.9001 \\
\bottomrule
\end{tabular}%
}
\caption{Coptic (cop) \dev results using predicted NLP annotations.}
\label{tab:dev-cop-predicted}
\end{table*}

\begin{table*}[tbp]
\centering
\resizebox{\textwidth}{!}{%
\begin{tabular}{llrrr}
\toprule
Setting & Model & BLEU & ChrF++ & BERTScore F1 \\
\midrule
Baseline & Gemma-4-31b & 0.9482 & 12.8595 & 0.8640 \\
Baseline & GPT-5.6-terra & 1.5441 & 12.8494 & 0.8622 \\
Baseline & Qwen3.5-27B & 0.8586 & 12.4159 & 0.8617 \\
Lexicon & Gemma-4-31b & 4.4488 & 21.5798 & 0.8856 \\
Lexicon & GPT-5.6-terra & 4.3653 & 21.4010 & 0.8834 \\
Lexicon & Qwen3.5-27B & 3.7827 & 20.6767 & 0.8804 \\
Lexicon + CoNLL-U & Gemma-4-31b & 6.7683 & 25.4041 & 0.8961 \\
Lexicon + CoNLL-U & GPT-5.6-terra & 8.2113 & 26.2428 & 0.8965 \\
Lexicon + CoNLL-U & Qwen3.5-27B & 6.0687 & 24.7066 & 0.8913 \\
Lexicon + Dependency & Gemma-4-31b & 4.7771 & 22.0941 & 0.8882 \\
Lexicon + Dependency & GPT-5.6-terra & 5.0422 & 22.2082 & 0.8865 \\
Lexicon + Dependency & Qwen3.5-27B & 4.0177 & 21.5110 & 0.8815 \\
Lexicon + CoNLL-U + Dependency & Gemma-4-31b & 6.4150 & 25.1450 & 0.8948 \\
Lexicon + CoNLL-U + Dependency & GPT-5.6-terra & 7.6730 & 26.2328 & 0.8957 \\
Lexicon + CoNLL-U + Dependency & Qwen3.5-27B & 6.1720 & 24.6569 & 0.8913 \\
\bottomrule
\end{tabular}%
}
\caption{Chintang (ctn) \dev results using predicted NLP annotations.}
\label{tab:dev-ctn-predicted}
\end{table*}

\begin{table*}[tbp]
\centering
\resizebox{\textwidth}{!}{%
\begin{tabular}{llrrr}
\toprule
Setting & Model & BLEU & ChrF++ & BERTScore F1 \\
\midrule
Baseline & Gemma-4-31b & 0.9776 & 19.1242 & 0.8525 \\
Baseline & GPT-5.6-terra & 5.9578 & 26.6827 & 0.8809 \\
Baseline & Qwen3.5-27B & 1.1914 & 19.1487 & 0.8449 \\
Lexicon & Gemma-4-31b & 1.1382 & 20.5770 & 0.8561 \\
Lexicon & GPT-5.6-terra & 5.8671 & 27.6324 & 0.8837 \\
Lexicon & Qwen3.5-27B & 1.6406 & 20.6398 & 0.8521 \\
Lexicon + CoNLL-U & Gemma-4-31b & 1.9301 & 21.4229 & 0.8655 \\
Lexicon + CoNLL-U & GPT-5.6-terra & 6.4252 & 28.6647 & 0.8849 \\
Lexicon + CoNLL-U & Qwen3.5-27B & 2.7482 & 22.6021 & 0.8634 \\
Lexicon + Dependency & Gemma-4-31b & 2.0017 & 21.4267 & 0.8619 \\
Lexicon + Dependency & GPT-5.6-terra & 5.4007 & 27.2167 & 0.8817 \\
Lexicon + Dependency & Qwen3.5-27B & 2.3913 & 21.6105 & 0.8597 \\
Lexicon + CoNLL-U + Dependency & Gemma-4-31b & 2.1558 & 22.2045 & 0.8658 \\
Lexicon + CoNLL-U + Dependency & GPT-5.6-terra & 6.1967 & 28.1970 & 0.8832 \\
Lexicon + CoNLL-U + Dependency & Qwen3.5-27B & 2.6442 & 22.2967 & 0.8640 \\
\bottomrule
\end{tabular}%
}
\caption{Xibe (sjo) \dev results using predicted NLP annotations.}
\label{tab:dev-sjo-predicted}
\end{table*}

\paragraph{Test: Gold}
\label{sec:app-gold-results}

Full metrics on the \texttt{test} set using gold data can be found in Tables \ref{tab:bej-gold}, \ref{tab:ckt-gold}, \ref{tab:cop-gold}, \ref{tab:ctn-gold}, and \ref{tab:sjo-gold}.

\begin{table*}[tbp]
\centering
\resizebox{\textwidth}{!}{%
\begin{tabular}{llrrr}
\toprule
Setting & Model & BLEU & ChrF++ & BERTScore F1 \\
\midrule
Baseline & Gemma-4-31b & 0.3166 & 13.5255 & 0.8554 \\
Baseline & GPT-5.6-terra & 0.4430 & 13.5159 & 0.8590 \\
Baseline & Qwen3.5-27B & 0.3664 & 13.5176 & 0.8494 \\
Lexicon & Gemma-4-31b & 1.3116 & 20.0209 & 0.8702 \\
Lexicon & GPT-5.6-terra & 1.2960 & 19.1972 & 0.8716 \\
Lexicon & Qwen3.5-27B & 1.4384 & 20.2500 & 0.8654 \\
Lexicon + CoNLL-U & Gemma-4-31b & 3.3184 & 24.2778 & 0.8786 \\
Lexicon + CoNLL-U & GPT-5.6-terra & 3.3547 & 23.6862 & 0.8804 \\
Lexicon + CoNLL-U & Qwen3.5-27B & 2.4043 & 23.0962 & 0.8736 \\
Lexicon + Dependency & Gemma-4-31b & 2.2356 & 21.0515 & 0.8743 \\
Lexicon + Dependency & GPT-5.6-terra & 2.7090 & 21.0722 & 0.8780 \\
Lexicon + Dependency & Qwen3.5-27B & 1.6107 & 20.0803 & 0.8711 \\
Lexicon + CoNLL-U + Dependency & Gemma-4-31b & 3.6357 & 24.1889 & 0.8799 \\
Lexicon + CoNLL-U + Dependency & GPT-5.6-terra & 3.4821 & 23.7514 & 0.8816 \\
Lexicon + CoNLL-U + Dependency & Qwen3.5-27B & 2.7867 & 23.2770 & 0.8743 \\
\bottomrule
\end{tabular}%
}
\caption{Beja (bej) \texttt{test} results using the gold treebank.}
\label{tab:bej-gold}
\end{table*}

\begin{table*}[tbp]
\centering
\resizebox{\textwidth}{!}{%
\begin{tabular}{llrrr}
\toprule
Setting & Model & BLEU & ChrF++ & BERTScore F1 \\
\midrule
Baseline & Gemma-4-31b & 0.3560 & 13.4491 & 0.8612 \\
Baseline & GPT-5.6-terra & 0.4862 & 13.4033 & 0.8659 \\
Baseline & Qwen3.5-27B & 0.7019 & 13.4046 & 0.8572 \\
Lexicon & Gemma-4-31b & 0.9594 & 15.3322 & 0.8640 \\
Lexicon & GPT-5.6-terra & 0.6871 & 14.2985 & 0.8676 \\
Lexicon & Qwen3.5-27B & 0.8670 & 14.1371 & 0.8599 \\
Lexicon + CoNLL-U & Gemma-4-31b & 1.3466 & 15.8990 & 0.8690 \\
Lexicon + CoNLL-U & GPT-5.6-terra & 1.9115 & 16.2252 & 0.8757 \\
Lexicon + CoNLL-U & Qwen3.5-27B & 1.1628 & 14.5690 & 0.8660 \\
Lexicon + Dependency & Gemma-4-31b & 1.3860 & 15.2724 & 0.8671 \\
Lexicon + Dependency & GPT-5.6-terra & 2.1155 & 16.0821 & 0.8747 \\
Lexicon + Dependency & Qwen3.5-27B & 1.2099 & 14.5299 & 0.8662 \\
Lexicon + CoNLL-U + Dependency & Gemma-4-31b & 1.0199 & 15.9192 & 0.8690 \\
Lexicon + CoNLL-U + Dependency & GPT-5.6-terra & 1.2782 & 15.7662 & 0.8752 \\
Lexicon + CoNLL-U + Dependency & Qwen3.5-27B & 1.1002 & 14.8257 & 0.8680 \\
\bottomrule
\end{tabular}%
}
\caption{Chukchi (ckt) \texttt{test} results using the gold treebank.}
\label{tab:ckt-gold}
\end{table*}

\begin{table*}[tbp]
\centering
\resizebox{\textwidth}{!}{%
\begin{tabular}{llrrr}
\toprule
Setting & Model & BLEU & ChrF++ & BERTScore F1 \\
\midrule
Baseline & Gemma-4-31b & 6.4323 & 26.6546 & 0.8718 \\
Baseline & GPT-5.6-terra & 31.2051 & 52.4880 & 0.9314 \\
Baseline & Qwen3.5-27B & 2.4084 & 19.8988 & 0.8515 \\
Lexicon & Gemma-4-31b & 13.3405 & 37.3074 & 0.8972 \\
Lexicon & GPT-5.6-terra & 31.3266 & 53.1647 & 0.9338 \\
Lexicon & Qwen3.5-27B & 10.8785 & 33.1268 & 0.8857 \\
Lexicon + CoNLL-U & Gemma-4-31b & 17.0616 & 41.7160 & 0.9084 \\
Lexicon + CoNLL-U & GPT-5.6-terra & 31.6083 & 53.5560 & 0.9354 \\
Lexicon + CoNLL-U & Qwen3.5-27B & 16.3000 & 40.4149 & 0.9034 \\
Lexicon + Dependency & Gemma-4-31b & 15.8184 & 39.9953 & 0.9045 \\
Lexicon + Dependency & GPT-5.6-terra & 31.6076 & 53.3641 & 0.9342 \\
Lexicon + Dependency & Qwen3.5-27B & 13.2884 & 36.4747 & 0.8965 \\
Lexicon + CoNLL-U + Dependency & Gemma-4-31b & 17.5501 & 42.0839 & 0.9090 \\
Lexicon + CoNLL-U + Dependency & GPT-5.6-terra & 31.3905 & 53.3146 & 0.9346 \\
Lexicon + CoNLL-U + Dependency & Qwen3.5-27B & 16.7075 & 40.9514 & 0.9053 \\
\bottomrule
\end{tabular}%
}
\caption{Coptic (cop) texttt{test} results using the gold treebank.}
\label{tab:cop-gold}
\end{table*}

\begin{table*}[tbp]
\centering
\resizebox{\textwidth}{!}{%
\begin{tabular}{llrrr}
\toprule
Setting & Model & BLEU & ChrF++ & BERTScore F1 \\
\midrule
Baseline & Gemma-4-31b & 0.7070 & 13.1166 & 0.8664 \\
Baseline & GPT-5.6-terra & 1.4518 & 13.0424 & 0.8657 \\
Baseline & Qwen3.5-27B & 1.0067 & 12.5322 & 0.8630 \\
Lexicon & Gemma-4-31b & 4.6571 & 21.8816 & 0.8900 \\
Lexicon & GPT-5.6-terra & 4.8112 & 22.4292 & 0.8895 \\
Lexicon & Qwen3.5-27B & 3.8300 & 21.3008 & 0.8856 \\
Lexicon + CoNLL-U & Gemma-4-31b & 8.3491 & 28.1169 & 0.9036 \\
Lexicon + CoNLL-U & GPT-5.6-terra & 10.4980 & 30.5123 & 0.9085 \\
Lexicon + CoNLL-U & Qwen3.5-27B & 7.7245 & 26.9743 & 0.9004 \\
Lexicon + Dependency & Gemma-4-31b & 5.2018 & 23.5193 & 0.8953 \\
Lexicon + Dependency & GPT-5.6-terra & 6.1520 & 24.4227 & 0.8958 \\
Lexicon + Dependency & Qwen3.5-27B & 4.3612 & 22.6643 & 0.8906 \\
Lexicon + CoNLL-U + Dependency & Gemma-4-31b & 8.5990 & 28.6246 & 0.9050 \\
Lexicon + CoNLL-U + Dependency & GPT-5.6-terra & 10.9109 & 30.7102 & 0.9086 \\
Lexicon + CoNLL-U + Dependency & Qwen3.5-27B & 7.9721 & 27.3312 & 0.9011 \\
\bottomrule
\end{tabular}%
}
\caption{Chintang (ctn) \texttt{test} results using the gold treebank.}
\label{tab:ctn-gold}
\end{table*}

\begin{table*}[tbp]
\centering
\resizebox{\textwidth}{!}{%
\begin{tabular}{llrrr}
\toprule
Setting & Model & BLEU & ChrF++ & BERTScore F1 \\
\midrule
Baseline & Gemma-4-31b & 1.0558 & 19.3262 & 0.8570 \\
Baseline & GPT-5.6-terra & 4.9633 & 26.2467 & 0.8833 \\
Baseline & Qwen3.5-27B & 1.3038 & 19.5234 & 0.8488 \\
Lexicon & Gemma-4-31b & 1.3428 & 19.5705 & 0.8605 \\
Lexicon & GPT-5.6-terra & 5.2909 & 26.8043 & 0.8863 \\
Lexicon & Qwen3.5-27B & 1.0025 & 19.8530 & 0.8538 \\
Lexicon + CoNLL-U & Gemma-4-31b & 2.8933 & 23.0188 & 0.8753 \\
Lexicon + CoNLL-U & GPT-5.6-terra & 7.4236 & 29.1279 & 0.8935 \\
Lexicon + CoNLL-U & Qwen3.5-27B & 3.2601 & 23.5492 & 0.8732 \\
Lexicon + Dependency & Gemma-4-31b & 2.0846 & 20.8511 & 0.8660 \\
Lexicon + Dependency & GPT-5.6-terra & 5.2165 & 27.5223 & 0.8888 \\
Lexicon + Dependency & Qwen3.5-27B & 2.4510 & 21.8197 & 0.8662 \\
Lexicon + CoNLL-U + Dependency & Gemma-4-31b & 3.1436 & 22.8753 & 0.8752 \\
Lexicon + CoNLL-U + Dependency & GPT-5.6-terra & 7.3409 & 30.2342 & 0.8964 \\
Lexicon + CoNLL-U + Dependency & Qwen3.5-27B & 2.9798 & 23.3940 & 0.8735 \\
\bottomrule
\end{tabular}%
}
\caption{Xibe (sjo) \texttt{test} results using the gold treebank.}
\label{tab:sjo-gold}
\end{table*}

\paragraph{Test: Predicted NLP}
\label{sec:app-pred-results}
Full metrics on the \texttt{test} set using predicted data can be found in Tables \ref{tab:bej-predicted}, \ref{tab:ckt-predicted}, \ref{tab:cop-predicted}, \ref{tab:ctn-predicted}, and \ref{tab:sjo-predicted}.

\begin{table*}[tbp]
\centering
\resizebox{\textwidth}{!}{%
\begin{tabular}{llrrr}
\toprule
Setting & Model & BLEU & ChrF++ & BERTScore F1 \\
\midrule
Baseline & Gemma-4-31b & 0.3869 & 13.2931 & 0.8541 \\
Baseline & GPT-5.6-terra & 0.3629 & 13.3222 & 0.8576 \\
Baseline & Qwen3.5-27B & 0.2762 & 12.8097 & 0.8493 \\
Lexicon & Gemma-4-31b & 1.3062 & 20.3689 & 0.8696 \\
Lexicon & GPT-5.6-terra & 1.3955 & 19.6067 & 0.8708 \\
Lexicon & Qwen3.5-27B & 1.4367 & 20.2212 & 0.8651 \\
Lexicon + CoNLL-U & Gemma-4-31b & 2.7921 & 23.4339 & 0.8767 \\
Lexicon + CoNLL-U & GPT-5.6-terra & 2.5983 & 23.1692 & 0.8771 \\
Lexicon + CoNLL-U & Qwen3.5-27B & 2.7807 & 22.8357 & 0.8722 \\
Lexicon + Dependency & Gemma-4-31b & 1.7644 & 20.5924 & 0.8743 \\
Lexicon + Dependency & GPT-5.6-terra & 2.1110 & 20.0839 & 0.8758 \\
Lexicon + Dependency & Qwen3.5-27B & 0.9691 & 20.2860 & 0.8710 \\
Lexicon + CoNLL-U + Dependency & Gemma-4-31b & 2.9696 & 23.3529 & 0.8767 \\
Lexicon + CoNLL-U + Dependency & GPT-5.6-terra & 2.7467 & 22.7453 & 0.8782 \\
Lexicon + CoNLL-U + Dependency & Qwen3.5-27B & 2.3524 & 22.4466 & 0.8733 \\
\bottomrule
\end{tabular}%
}
\caption{Beja (bej) \texttt{test} results using predicted NLP annotations.}
\label{tab:bej-predicted}
\end{table*}

\begin{table*}[tbp]
\centering
\resizebox{\textwidth}{!}{%
\begin{tabular}{llrrr}
\toprule
Setting & Model & BLEU & ChrF++ & BERTScore F1 \\
\midrule
Baseline & Gemma-4-31b & 0.4125 & 14.2330 & 0.8608 \\
Baseline & GPT-5.6-terra & 0.5480 & 13.2287 & 0.8643 \\
Baseline & Qwen3.5-27B & 0.6886 & 13.6135 & 0.8573 \\
Lexicon & Gemma-4-31b & 0.5775 & 14.9631 & 0.8626 \\
Lexicon & GPT-5.6-terra & 0.9376 & 14.5231 & 0.8668 \\
Lexicon & Qwen3.5-27B & 1.1107 & 14.2552 & 0.8597 \\
Lexicon + CoNLL-U & Gemma-4-31b & 0.7975 & 14.6903 & 0.8647 \\
Lexicon + CoNLL-U & GPT-5.6-terra & 1.1758 & 14.7222 & 0.8702 \\
Lexicon + CoNLL-U & Qwen3.5-27B & 1.0226 & 13.9235 & 0.8631 \\
Lexicon + Dependency & Gemma-4-31b & 1.2640 & 14.7414 & 0.8640 \\
Lexicon + Dependency & GPT-5.6-terra & 1.5119 & 14.8541 & 0.8708 \\
Lexicon + Dependency & Qwen3.5-27B & 1.2006 & 14.4031 & 0.8642 \\
Lexicon + CoNLL-U + Dependency & Gemma-4-31b & 0.6936 & 14.9919 & 0.8661 \\
Lexicon + CoNLL-U + Dependency & GPT-5.6-terra & 1.1761 & 14.7298 & 0.8712 \\
Lexicon + CoNLL-U + Dependency & Qwen3.5-27B & 0.7800 & 13.4409 & 0.8640 \\
\bottomrule
\end{tabular}%
}
\caption{Chukchi (ckt) \texttt{test} results using predicted NLP annotations.}
\label{tab:ckt-predicted}
\end{table*}

\begin{table*}[tbp]
\centering
\resizebox{\textwidth}{!}{%
\begin{tabular}{llrrr}
\toprule
Setting & Model & BLEU & ChrF++ & BERTScore F1 \\
\midrule
Baseline & Gemma-4-31b & 6.3062 & 26.6146 & 0.8708 \\
Baseline & GPT-5.6-terra & 30.8087 & 52.3420 & 0.9316 \\
Baseline & Qwen3.5-27B & 2.7540 & 19.9402 & 0.8514 \\
Lexicon & Gemma-4-31b & 13.9570 & 37.5141 & 0.8978 \\
Lexicon & GPT-5.6-terra & 30.7845 & 52.9620 & 0.9335 \\
Lexicon & Qwen3.5-27B & 10.5563 & 32.5239 & 0.8848 \\
Lexicon + CoNLL-U & Gemma-4-31b & 16.5128 & 41.0786 & 0.9059 \\
Lexicon + CoNLL-U & GPT-5.6-terra & 31.6931 & 53.6593 & 0.9352 \\
Lexicon + CoNLL-U & Qwen3.5-27B & 15.4194 & 39.5010 & 0.9006 \\
Lexicon + Dependency & Gemma-4-31b & 15.1499 & 39.1182 & 0.9024 \\
Lexicon + Dependency & GPT-5.6-terra & 31.3243 & 53.1205 & 0.9342 \\
Lexicon + Dependency & Qwen3.5-27B & 12.5531 & 34.8885 & 0.8918 \\
Lexicon + CoNLL-U + Dependency & Gemma-4-31b & 16.5170 & 41.1750 & 0.9061 \\
Lexicon + CoNLL-U + Dependency & GPT-5.6-terra & 31.1249 & 53.1993 & 0.9341 \\
Lexicon + CoNLL-U + Dependency & Qwen3.5-27B & 15.3499 & 39.3763 & 0.9000 \\
\bottomrule
\end{tabular}%
}
\caption{Coptic (cop) \texttt{test} results using predicted NLP annotations.}
\label{tab:cop-predicted}
\end{table*}

\begin{table*}[tbp]
\centering
\resizebox{\textwidth}{!}{%
\begin{tabular}{llrrr}
\toprule
Setting & Model & BLEU & ChrF++ & BERTScore F1 \\
\midrule
Baseline & Gemma-4-31b & 0.6314 & 12.0142 & 0.8654 \\
Baseline & GPT-5.6-terra & 1.2025 & 13.1445 & 0.8652 \\
Baseline & Qwen3.5-27B & 1.0251 & 12.6076 & 0.8633 \\
Lexicon & Gemma-4-31b & 4.3077 & 21.7345 & 0.8892 \\
Lexicon & GPT-5.6-terra & 4.7300 & 21.9831 & 0.8886 \\
Lexicon & Qwen3.5-27B & 3.9613 & 21.3352 & 0.8864 \\
Lexicon + CoNLL-U & Gemma-4-31b & 7.3122 & 26.5200 & 0.9004 \\
Lexicon + CoNLL-U & GPT-5.6-terra & 8.4274 & 27.8454 & 0.9029 \\
Lexicon + CoNLL-U & Qwen3.5-27B & 6.7252 & 25.5427 & 0.8973 \\
Lexicon + Dependency & Gemma-4-31b & 4.5125 & 22.3806 & 0.8924 \\
Lexicon + Dependency & GPT-5.6-terra & 5.8426 & 23.2356 & 0.8922 \\
Lexicon + Dependency & Qwen3.5-27B & 3.9788 & 21.6789 & 0.8886 \\
Lexicon + CoNLL-U + Dependency & Gemma-4-31b & 6.7909 & 26.1376 & 0.9000 \\
Lexicon + CoNLL-U + Dependency & GPT-5.6-terra & 7.3258 & 27.0831 & 0.9011 \\
Lexicon + CoNLL-U + Dependency & Qwen3.5-27B & 7.2938 & 25.9208 & 0.8978 \\
\bottomrule
\end{tabular}
}
\caption{Chintang (ctn) \texttt{test} results using predicted NLP annotations.}
\label{tab:ctn-predicted}
\end{table*}

\begin{table*}[tbp]
\centering
\resizebox{\textwidth}{!}{%
\begin{tabular}{llrrr}
\toprule
Setting & Model & BLEU & ChrF++ & BERTScore F1 \\
\midrule
Baseline & Gemma-4-31b & 1.1051 & 19.0809 & 0.8552 \\
Baseline & GPT-5.6-terra & 4.6174 & 26.1733 & 0.8824 \\
Baseline & Qwen3.5-27B & 0.8305 & 19.0194 & 0.8497 \\
Lexicon & Gemma-4-31b & 1.4615 & 20.3558 & 0.8606 \\
Lexicon & GPT-5.6-terra & 5.6454 & 27.1608 & 0.8867 \\
Lexicon & Qwen3.5-27B & 1.1249 & 19.5336 & 0.8538 \\
Lexicon + CoNLL-U & Gemma-4-31b & 1.6298 & 20.9510 & 0.8695 \\
Lexicon + CoNLL-U & GPT-5.6-terra & 7.0767 & 29.0004 & 0.8911 \\
Lexicon + CoNLL-U & Qwen3.5-27B & 2.6840 & 22.3354 & 0.8693 \\
Lexicon + Dependency & Gemma-4-31b & 1.4133 & 20.3458 & 0.8640 \\
Lexicon + Dependency & GPT-5.6-terra & 5.4325 & 27.9321 & 0.8865 \\
Lexicon + Dependency & Qwen3.5-27B & 2.3163 & 20.4067 & 0.8654 \\
Lexicon + CoNLL-U + Dependency & Gemma-4-31b & 2.1081 & 21.6094 & 0.8705 \\
Lexicon + CoNLL-U + Dependency & GPT-5.6-terra & 5.7781 & 27.9827 & 0.8877 \\
Lexicon + CoNLL-U + Dependency & Qwen3.5-27B & 2.6542 & 22.1389 & 0.8698 \\
\bottomrule
\end{tabular}%
}
\caption{Xibe (sjo) \texttt{test} results using predicted NLP annotations.}
\label{tab:sjo-predicted}
\end{table*}

\subsection{Effect of length}
\label{sec:app-gold-length}
\begin{figure*}[t]
    \centering
    \includegraphics[width=\textwidth]{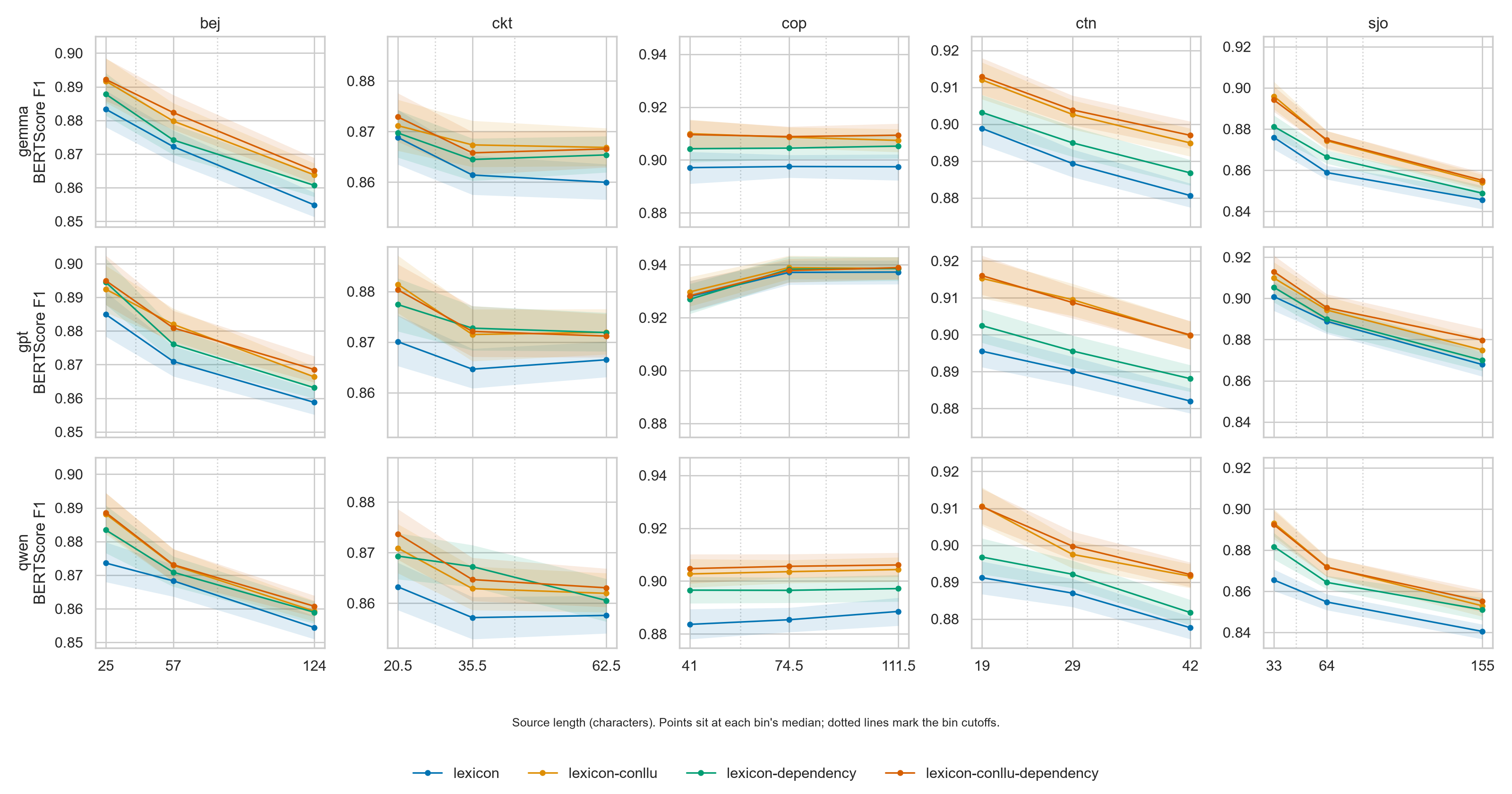}
    \caption{BERTScore F1 plotted against source length, measured in characters. Rows are models and columns are languages. Sentences are divided into three length bins within each language. Dotted vertical lines mark the bin cutoffs; colored lines show prompt settings; shaded bands show 95\% confidence intervals. }
    \label{fig:gold_char}
\end{figure*}

\begin{figure*}[t]
    \centering
    \includegraphics[width=\textwidth]{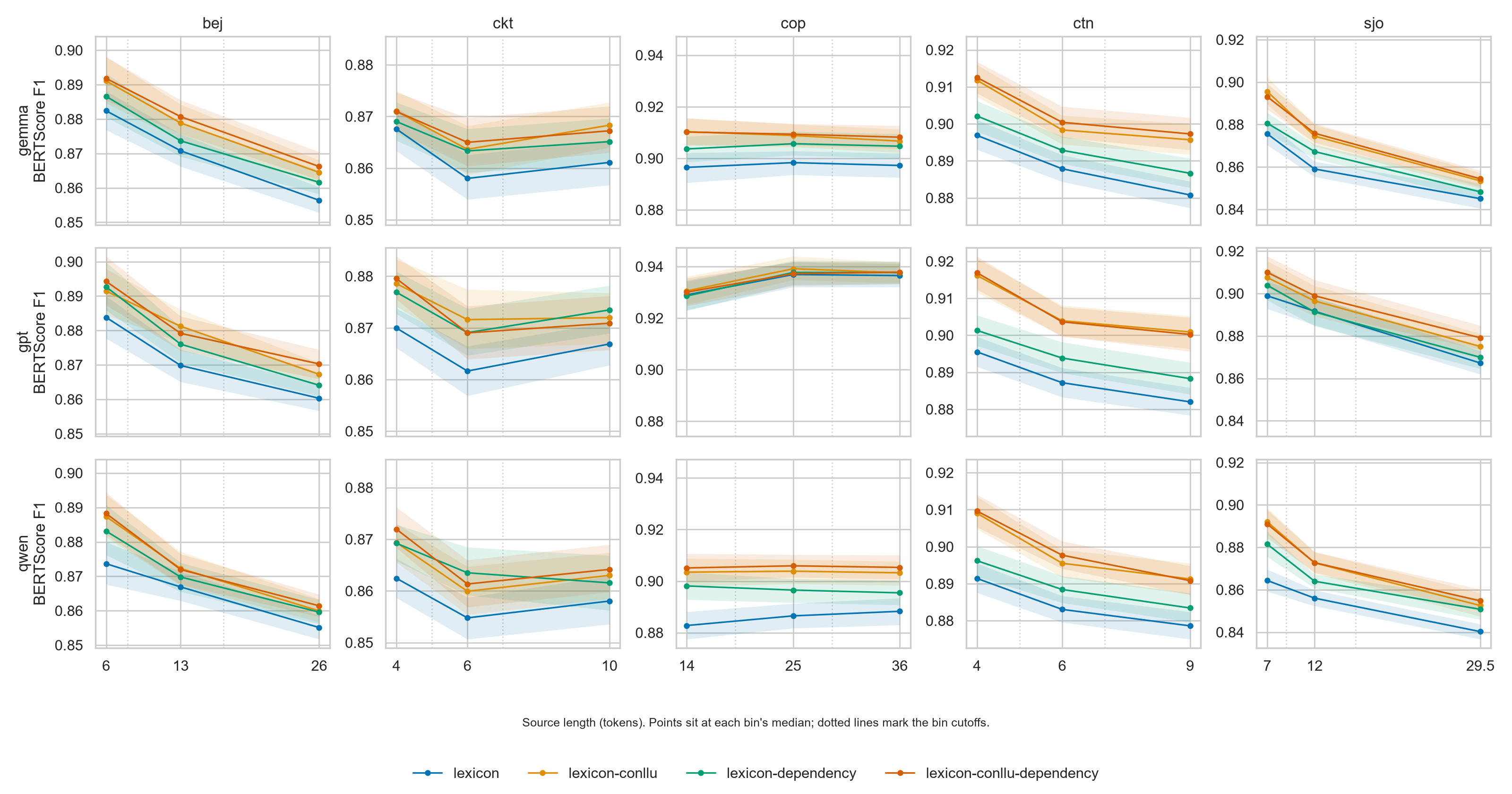}
    \caption{BERTScore F1 plotted against source length, measured in tokens. Rows are models and columns are languages. Sentences are divided into three length bins within each language. Dotted vertical lines mark the bin cutoffs; colored lines show prompt settings; shaded bands show 95\% confidence intervals. }
    \label{fig:gold_tok}
\end{figure*}

\paragraph{Binning}
For each language, we split the source sentences into three bins (terciles) using that language's own length distribution, so the cutoffs differ across languages. The same sentences are translated by all models, so the bins are shared across models. Each plotted point is the mean BERTScore F1 over the sentences in a bin, placed at the bin's median length. Shaded bands are 95\% bootstrap confidence intervals of that mean. Because bins are computed within each language, the axes not comparable across languages. Figure \ref{fig:gold_char} displays the trends when the measure of length is a character; Figure \ref{fig:gold_tok} shows the same when the measure of length is a token. 

\subsection{Examples}
\label{sec:app-examples}
\begin{table*}[!ht]
\centering
\small
\setlength{\tabcolsep}{4pt}
\renewcommand{\arraystretch}{1.1}

\begin{tabular}{@{}l p{0.43\textwidth} r r r@{}}
\toprule
\textbf{Setting} &
\textbf{Translation, Chintang to English} &
\textbf{$\Delta$ BLEU} &
\textbf{$\Delta$ chrF++} &
\textbf{$\Delta$ F1} \\
\midrule

\textbf{Gemma-4-31B} & & & & \\

Reference
& \textit{I live up there, on top of the ravine.}
& -- & -- & -- \\

Baseline
& \textit{I have a sister, Khojabhandu.}
& 0 & 0 & 0 \\

\lexicon
& \textit{That is where the ravine is.}
& +3.3277 & +17.8306 & +0.0587 \\

\all
& \textit{Oh, I live up there in the ravine.}
& +30.6625 & +52.4039 & +0.1023 \\[3pt]

\cmidrule(lr){1-5}

Reference
& \textit{Anita introduced Lokendra and Lakhman to each other.}
& -- & -- & -- \\

Baseline
& \textit{Anita forgot Lokendran's Lakhman's eyes.}
& 0 & 0 & 0 \\

\lexicon
& \textit{Anita knows Lokendra and Lakhman.}
& +14.9674 & +17.2812 & +0.0509 \\

\all
& \textit{Anita introduced Lokendra to Lakhman.}
& +15.9384 & +31.4992 & +0.0771 \\

\midrule

\textbf{Qwen3.5-27B} & & & & \\

Reference
& \textit{Kalpana pulls out people's hair!}
& -- & -- & -- \\

Baseline
& \textit{Imagine that you are a bird flying in the sky!}
& 0 & 0 & 0 \\

\lexicon
& \textit{Pull out Kalpana's hair!}
& +15.6099 & +32.4356 & +0.0599 \\

\all
& \textit{Kalpana is pulling out my hair!}
& +11.8719 & +28.9687 & +0.1238 \\[3pt]

\cmidrule(lr){1-5}

Reference
& \textit{I slid down from the terrace.}
& -- & -- & -- \\

Baseline
& \textit{I am going to the forest.}
& 0 & 0 & 0 \\

\lexicon
& \textit{The terrace slipped.}
& +0.3300 & +16.3617 & +0.0191 \\

\all
& \textit{I slipped on the terrace.}
& +18.8395 & +33.9077 & +0.0471 \\

\midrule

\textbf{GPT-5.6 Terra} & & & & \\

Reference
& \textit{There is no money to buy it, they won't give it to us.}
& -- & -- & -- \\

Baseline
& \textit{Why did you do that, dear?}
& 0 & 0 & 0 \\

\lexicon
& \textit{Buy money, give me your buttocks.}
& +0.9536 & +12.7776 & +0.0133 \\

\all
& \textit{There is no money to buy it; they won't give us any.}
& +42.7756 & +58.0250 & +0.1213 \\[3pt]

\cmidrule(lr){1-5}

Reference
& \textit{It's said they didn't speak for a while.}
& -- & -- & -- \\

Baseline
& \textit{I miss you very much.}
& 0 & 0 & 0 \\

\lexicon
& \textit{Please speak for a moment.}
& +14.7691 & +23.8900 & +0.0189 \\

\all
& \textit{They reportedly did not speak for a while.}
& +39.7093 & +48.2364 & +0.0969 \\

\bottomrule
\end{tabular}

\caption{Sample \texttt{dev} outputs for Chintang with Gemma-4-31B,
Qwen3.5-27B, and GPT-5.6 Terra under each setting. Metric improvements
$\Delta$ are reported relative to the baseline for BLEU, chrF++, and
BERTScore F1.}
\label{tab:err-ana}
\vspace{-5pt}
\end{table*}

\begin{table*}[!ht]
\centering
\small
\setlength{\tabcolsep}{4pt}
\renewcommand{\arraystretch}{1.12}

\begin{tabular}{@{}l p{0.49\textwidth} r r r@{}}
\toprule
\textbf{Setting} &
\textbf{Translation, Source to English} &
\textbf{$\Delta$ BLEU} &
\textbf{$\Delta$ chrF++} &
\textbf{$\Delta$ F1} \\
\midrule

\textbf{Beja} & & & & \\

Reference
& \textit{They wanted to eat him.}
& --
& --
& -- \\

Baseline
& \textit{The big house is here.}
& 0
& 0
& 0 \\

\lexicon
& \textit{Those are the big ones.}
& 0.0000
& +4.0334
& $-$0.0018 \\

\all
& \textit{They wanted to kill him.}
& +14.8408
& +49.9049
& +0.0967 \\[3pt]

\cmidrule(lr){1-5}

Reference
& \textit{His children and himself were really happy.}
& --
& --
& -- \\

Baseline
& \textit{The rain fell, the crops grew, and the land became green.}
& 0
& 0
& 0 \\

\lexicon
& \textit{I hid it, and I was happy that it was found.}
& +0.6883
& +1.5832
& +0.0188 \\

\all
& \textit{Their relatives and he were happy.}
& +10.1632
& +12.9169
& +0.0838 \\

\midrule

\textbf{Coptic} & & & & \\

Reference
& \textit{They are like the light of day.}
& --
& --
& -- \\

Baseline
& \textit{And they went to their works that day.}
& 0
& 0
& 0 \\

\lexicon
& \textit{And they likened it to the light of the day.}
& +10.2234
& +29.6470
& +0.0681 \\

\all
& \textit{And they were like the light of the day.}
& +23.0085
& +48.3562
& +0.0898 \\[3pt]

\cmidrule(lr){1-5}

Reference
& \textit{He also said that Abba Pambo said: ``This is the way that it is fitting for the monk to wear his clothes:}
& --
& --
& -- \\

Baseline
& \textit{And he said, ``Father, my lord, this is the place that is desired by the monks, for they are dwelling there.''}
& 0
& 0
& 0 \\

\lexicon
& \textit{He again said, ``Father, my abbot, tell me what is the proper manner for a monk to wear his garments.''}
& +8.2491
& +2.7772
& +0.0216 \\

\all
& \textit{He again said that Abbot Ambow said, ``This is the manner that is fitting for a monk to wear his garments.''}
& +13.9876
& +25.6272
& +0.0665 \\

\midrule

\textbf{Xibe} & & & & \\

Reference
& \textit{Let us go sightseeing to the Ili River!}
& --
& --
& -- \\

Baseline
& \textit{The yellow flowers bloom in the spring, and the wind blows gently!}
& 0
& 0
& 0 \\

\lexicon
& \textit{The yellow flowers bloom again in the spring.}
& +1.3905
& $-$2.2396
& +0.0012 \\

\all
& \textit{Let us go to the valley of the Elbi river!}
& +14.9831
& +24.6071
& +0.0896 \\[3pt]

\cmidrule(lr){1-5}

Reference
& \textit{Sir, if you do not go out today, please come to my house to eat the dinner.}
& --
& --
& -- \\

Baseline
& \textit{Grandfather, I have already finished my homework; I'm just going to read a bit more before I go to sleep.}
& 0
& 0
& 0 \\

\lexicon
& \textit{Sir, you have come to this place for a reason; please tell me everything.}
& +3.6591
& +7.4293
& +0.0202 \\

\all
& \textit{Sir, if you do not go to the city, please come to my house for a visit.}
& +52.9928
& +44.7628
& +0.0864 \\

\midrule

\textbf{Chukchi} & & & & \\

Reference
& \textit{\guillemotleft This is me.}
& --
& --
& -- \\

Baseline
& \textit{I am a little bit tired.}
& 0
& 0
& 0 \\

\lexicon
& \textit{I am a little bit tired.}
& 0.0000
& 0.0000
& 0.0000 \\

\all
& \textit{It is I.}
& +12.4286
& +3.9484
& +0.0482 \\[3pt]

\cmidrule(lr){1-5}

Reference
& \textit{That is \guillemotleft Mrenti\guillemotright.}
& --
& --
& -- \\

Baseline
& \textit{I'm just waiting.}
& 0
& 0
& 0 \\

\lexicon
& \textit{The child is sleeping.}
& $-$3.2703
& +3.2864
& +0.0058 \\

\all
& \textit{It's Mrenti.}
& +3.7425
& +16.5310
& +0.0683 \\

\bottomrule
\end{tabular}

\caption{Sample \texttt{dev} outputs for Beja, Coptic, Xibe, and Chukchi with Gemma-4-31B under each setting. Metric improvements $\Delta$ are reported relative to the baseline for BLEU, chrF++, and BERTScore F1.}
\label{tab:err-ana-other}
\end{table*}

We include additional examples highlighting qualitative differences among settings, see \cref{tab:err-ana} and \cref{tab:err-ana-other}. The results show a range of improvements likely due to syntax parses (complex constructions), part-of-speech tags (e.g.~recognizing that `Mrenti' is a name in Chukchi, or a river name in Xibe) and morphological features (for example `It is I' in Chukchi).

\paragraph{Note on metrics and polysynthetic structure in Chukchi}
As shown in Table~\ref{tab:err-ana-other}, sentence-level BLEU improvements ($\Delta$\,BLEU) for Chukchi can appear small or non-intuitive on short sequences due to the strict $n$-gram exact matching requirements of BLEU \cite{post-2018-call}. Chukchi is a highly polysynthetic language with immense morphological type diversity in inputs. Because single words often encode entire multi-morpheme predicate structures, translation variants diverge significantly in string space, resulting in minimal exact 1-to-1 lexical overlap with gold references even when translations are semantically valid. This is less so for more isolating languages, in which 1:1 correspondences from lexicon glosses are more frequent, leading to higher scores on simple lexical overlap metrics such as BLEU.

\subsection{Significance testing}
\label{sec:appendix-sig}
For each of the 15 language/model pairs, the same sentences were translated under five settings (baseline, \lexicon, \lexicon+\conll, \lexicon+\dep, \lexicon+\syntax), so we treat sentence-level BERTScore F1 as a repeated measure. We compared settings with paired $t$-tests (10 setting comparisons per language/model pair, 150 in total) and applied the Holm correction across all 150 to control the overall false-positive rate at 5\%. Wilcoxon signed-rank tests gave the same decision in 145 of 150 comparisons. Table~\ref{tab:sig-main} shows the mean difference for each comparison, marked where significant after correction.

\begin{table*}[t]
\centering\footnotesize
\begin{tabular}{llccccccc}
\toprule
Model & Lang & \shortstack{\lexicon\\$-$ base} & \shortstack{\lexicon+\conll\\$-$ \lexicon} & \shortstack{\lexicon+\dep\\$-$ \lexicon} & \shortstack{\lexicon+\syntax\\$-$ \lexicon} & \shortstack{\lexicon+\conll\\$-$ \lexicon+\dep} & \shortstack{\lexicon+\syntax\\$-$ \lexicon+\conll} & \shortstack{\lexicon+\syntax\\$-$ \lexicon+\dep} \\
\midrule
Gemma & bej & +1.49$^{*}$ & +0.83$^{*}$ & +0.41$^{*}$ & +0.97$^{*}$ & +0.42$^{*}$ & +0.14\hphantom{$^{*}$} & +0.56$^{*}$ \\
Gemma & ckt & +0.29\hphantom{$^{*}$} & +0.50$^{*}$ & +0.31\hphantom{$^{*}$} & +0.50$^{*}$ & +0.19\hphantom{$^{*}$} & +0.00\hphantom{$^{*}$} & +0.19\hphantom{$^{*}$} \\
Gemma & cop & +2.56$^{*}$ & +1.13$^{*}$ & +0.73$^{*}$ & +1.19$^{*}$ & +0.40$^{*}$ & +0.06\hphantom{$^{*}$} & +0.46$^{*}$ \\
Gemma & ctn & +2.37$^{*}$ & +1.36$^{*}$ & +0.53$^{*}$ & +1.49$^{*}$ & +0.82$^{*}$ & +0.14\hphantom{$^{*}$} & +0.96$^{*}$ \\
Gemma & sjo & +0.34$^{*}$ & +1.48$^{*}$ & +0.54$^{*}$ & +1.46$^{*}$ & +0.94$^{*}$ & $-$0.02\hphantom{$^{*}$} & +0.92$^{*}$ \\
\midrule
GPT & bej & +1.26$^{*}$ & +0.87$^{*}$ & +0.64$^{*}$ & +0.99$^{*}$ & +0.23\hphantom{$^{*}$} & +0.12\hphantom{$^{*}$} & +0.36\hphantom{$^{*}$} \\
GPT & ckt & +0.17\hphantom{$^{*}$} & +0.80$^{*}$ & +0.70$^{*}$ & +0.76$^{*}$ & +0.10\hphantom{$^{*}$} & $-$0.04\hphantom{$^{*}$} & +0.06\hphantom{$^{*}$} \\
GPT & cop & +0.24\hphantom{$^{*}$} & +0.16\hphantom{$^{*}$} & +0.05\hphantom{$^{*}$} & +0.09\hphantom{$^{*}$} & +0.11\hphantom{$^{*}$} & $-$0.07\hphantom{$^{*}$} & +0.04\hphantom{$^{*}$} \\
GPT & ctn & +2.38$^{*}$ & +1.90$^{*}$ & +0.62$^{*}$ & +1.91$^{*}$ & +1.28$^{*}$ & +0.01\hphantom{$^{*}$} & +1.29$^{*}$ \\
GPT & sjo & +0.29\hphantom{$^{*}$} & +0.72$^{*}$ & +0.26\hphantom{$^{*}$} & +1.02$^{*}$ & +0.46\hphantom{$^{*}$} & +0.30\hphantom{$^{*}$} & +0.76$^{*}$ \\
\midrule
Qwen & bej & +1.61$^{*}$ & +0.81$^{*}$ & +0.57$^{*}$ & +0.88$^{*}$ & +0.24\hphantom{$^{*}$} & +0.07\hphantom{$^{*}$} & +0.31\hphantom{$^{*}$} \\
Qwen & ckt & +0.27$^{*}$ & +0.60$^{*}$ & +0.64$^{*}$ & +0.79$^{*}$ & $-$0.04\hphantom{$^{*}$} & +0.19\hphantom{$^{*}$} & +0.15\hphantom{$^{*}$} \\
Qwen & cop & +3.44$^{*}$ & +1.77$^{*}$ & +1.09$^{*}$ & +1.96$^{*}$ & +0.68$^{*}$ & +0.19\hphantom{$^{*}$} & +0.87$^{*}$ \\
Qwen & ctn & +2.26$^{*}$ & +1.48$^{*}$ & +0.50$^{*}$ & +1.56$^{*}$ & +0.98$^{*}$ & +0.08\hphantom{$^{*}$} & +1.06$^{*}$ \\
Qwen & sjo & +0.51$^{*}$ & +1.92$^{*}$ & +1.22$^{*}$ & +1.96$^{*}$ & +0.70$^{*}$ & +0.04\hphantom{$^{*}$} & +0.74$^{*}$ \\
\midrule
\multicolumn{2}{l}{Significant (of 15)} & 11 & 14 & 12 & 14 & 8 & 0 & 9 \\
\bottomrule
\end{tabular}
\caption{Mean paired difference in sentence-level BERTScore F1, in points on a 0--100 scale (F1 $\times$ 100). Each column is the score from first setting minus the score from the second setting; a positive entry means the first setting scored higher. $^{*}$ marks differences significant by paired $t$-test after Holm correction across all 150 comparisons ($\alpha=0.05$).}
\label{tab:sig-main}
\end{table*}

\section{AI Tool Use}We used \href{https://claude.ai}{Claude}, \href{https://claude.ai/code}{Claude Code}, and \href{https://gemini.google.com/app}{Gemini} when developing the code for this paper.

\end{document}